\documentclass[sn-apa]{sn-jnl}

\usepackage{amsmath,amsfonts,amssymb,amsthm,epsfig,url,epstopdf,array}
\usepackage{subcaption}
\usepackage{soul}
\usepackage{makecell}
\usepackage{longtable, pdflscape,booktabs}
\usepackage{caption}
\usepackage{csquotes} 
\usepackage{rotating}
\usepackage{hyperref}
\usepackage{graphicx}
\usepackage{float}
\usepackage{algorithm}
\usepackage{algpseudocode}
\usepackage{multirow}
\usepackage{color}

\makeatletter
\newenvironment{megaalgorithm}[1][h]{%
    \renewcommand{\ALG@name}{Pseudocode}
   \begin{algorithm}[#1]%
  }{\end{algorithm}}
\makeatother

\usepackage[normalem]{ulem} 
\usepackage[english]{babel}
\newcolumntype{C}[1]{>{\centering\arraybackslash}p{#1}}

\newcommand\MyBox[2]{
  \fbox{\lower0.75cm
    \vbox to 1.7cm{\vfil
      \hbox to 1.7cm{\hfil\parbox{1.4cm}{\centering #1\\#2}\hfil}
      \vfil}%
  }%
}

\jyear{2021}%

\theoremstyle{thmstyleone}%
\theoremstyle{thmstyletwo}%

\theoremstyle{thmstylethree}%

\begin{document}

\title[A survey on multi-objective HPO for ML]{A survey on multi-objective hyperparameter optimization algorithms for Machine Learning}


\author*[1,2]{\fnm{Alejandro} \sur{Morales-Hern\'andez}} \email{alejandro.moraleshernandez@uhasselt.be}

\author[1,2]{\fnm{Inneke} \sur{Van Nieuwenhuyse}}\email{inneke.vannieuwenhuyse@uhasselt.be}

\author[1,2,3]{\fnm{Sebastian} \sur{Rojas Gonzalez}}\email{sebastian.rojasgonzalez@uhasselt.be}

\affil[1]{\orgdiv{Faculty of Sciences}, \orgname{Hasselt University}, \orgaddress{\country{Belgium}}}

\affil[2]{\orgdiv{VCCM Core Lab and Data Science Institute}, \orgname{Hasselt University}, \orgaddress{\country{Belgium}}}

\affil[3]{\orgdiv{Surrogate Modeling Lab}, \orgname{Ghent University}, \orgaddress{\country{Belgium}}}


\abstract{Hyperparameter optimization (HPO) is a necessary step to ensure the best possible performance of Machine Learning (ML) algorithms. Several methods have been developed to perform HPO; most of these  are focused on optimizing one performance measure (usually an error-based measure), and the literature on such single-objective HPO problems is vast. Recently, though, algorithms have appeared that focus on optimizing multiple conflicting objectives simultaneously. This article presents a systematic survey of the literature published between 2014 and 2020 on multi-objective HPO algorithms, distinguishing between metaheuristic-based algorithms, metamodel-based algorithms and approaches using a mixture of both. We also discuss the quality metrics used to compare multi-objective HPO procedures and present future research directions.}




\maketitle

\section{Introduction}\label{sec1}

Nowadays, Artificial Intelligence (AI) is omnipresent in everyday life. Current technological advances allow us to analyze huge amounts of data to generate knowledge that is used in many different ways, e.g. for automatic user recommendations \citep{cai2020hybrid}, image recognition \citep{phillips2005overview,andreopoulos201350}, and supporting healthcare-related tasks \citep{jiang2017artificial}. In general, AI can be seen as a computer technology capable of carrying out functions that traditionally required human intelligence \citep{ertel2018introduction}. Although learning is a key element in many areas of artificial intelligence, the very concept of learning is mainly studied in the Machine Learning (ML) subfield. According to \cite{mitchell1997machine}, \textquote{a computer program is said to learn from experience \textbf{E} with respect to some class of tasks \textbf{T} and performance measure \textbf{P} if its performance at tasks in \textbf{T}, as measured by \textbf{P}, improves with experience \textbf{E}}. ML algorithms and their parameters must be intelligently configured to make the most of the data. Those parameters that need to be specified \emph{before} training the algorithm are usually referred to as \emph{hyperparameters}: they influence the learning process, but they are not optimized as part of the training algorithm.

The impact of these hyperparameters on algorithm performance should not be underestimated \citep{kim2017nemo,kong2017effect,singh2020classification,cooney2020evaluation}; yet, their optimization (hereafter referred to as \emph{hyperparameter optimization} or HPO) is a challenging task, as traditional optimization methods are often not applicable \citep{luo2016review}. Indeed, classic convex optimization methods such as gradient descent tend to be ill-suited for HPO, as the measure to optimize is usually a non-convex and non-differentiable function \citep{stamoulis2018designing, parsa2019pabo}. Furthermore, the hyperparameters to optimize may be discrete, categorical and/or continuous (typical hyperparameters for an Artificial Neural Network (ANN), for instance, are the number of layers, the number of neurons per layer, the type of optimizer, and the learning rate). The search space can also contain \emph{conditional hyperparameters}; e.g., the hyperparameters in a support vector machine algorithm depend on the type of kernel used. Finally, the time needed to train a machine learning model with a given hyperparameter configuration on a \emph{given} dataset may already be substantial, particularly for moderate to large datasets; as a common HPO algorithm requires multiple such training cycles, the algorithm itself needs to be computationally efficient to be useful in practice. 

HPO should not be confused with the more general topic of \emph{automatic algorithm configuration (AC)}, which is much broader in scope (see \cite{lopez2016irace, hutter2009paramils} for examples on this topic). In AC, in general, the aim is to find a well-performing parameter configuration for an arbitrary algorithm on a given, finite set of problem instances. In HPO, we typically search for a well-performing hyperparameter configuration on a \emph{single} data set, for a specific task (classification, image recognition, or other). 
The scope of AC is also broader than that of HPO, in the sense that the target algorithm does not necessarily carry out a learning process for the task under study; e.g., it also comprises the optimization of solvers and/or metaheuristics. 

HPO has gained increasing attention in recent years, probably spurred by the popularity of deep learning algorithms, which have demanding characteristics (e.g., the need for large amounts of data and time to train the models, high model complexity, and a diverse mix of hyperparameter types). Previously, analysts tended to use simple methods to look for the ``best'' hyperparameter settings. The most basic of these is grid search \citep{montgomery2017design}: the user creates a set of possible values for each hyperparameter, and the search evaluates the Cartesian product of these sets. Although this strategy is easy to implement and easy to understand, its performance is influenced by the number of hyperparameters to optimize, and the (number of) values chosen on the grid. Random search \citep{bergstra2012random} provides an alternative to grid search, and tends to be popular when some of the hyperparameters are more important than others; e.g, learning rate and momentum are critical to guarantee a faster convergence of neural networks \citep{guo2020deep}. More advanced optimization methods have also been put forward, such as meta-learning methods \citep{bui2020optimal}, neural architecture search (NAS) methods \citep{jing2020building}, multi-fidelity algorithms (such as Freeze-thaw Bayesian optimization \citep{swersky2014freeze}, Successive halving algorithm \citep{karnin2013almost}, Hyperband \citep{li2017hyperband}, Bayesian Optimization Hyperband \citep{falkner2018bohb}, and Multi-task Bayesian optimization \citep{swersky2013multi}), population-based optimization algorithms (such as Population-based training (PBT) \citep{jaderberg2017population} and Population-based Bandits (PB2) \citep{parker2020provably}), and reinforcement learning algorithms (such as HypRL \citep{jomaa2019hyp}) and the model-based Reinforcement Learning algorithm \citep{wu2020efficient}). 

So far, these more advanced approaches have largely focused on \emph{single-objective} HPO problems. \emph{Multi-objective} optimization is particularly relevant in HPO, as different conflicting objectives may be important for the analyst (e.g., the error-based performance of the target ML algorithm, inference time, model size, energy consumption, etc.). Multi-objective HPO should not be confused with multi-task learning (MTL). In multi-objective HPO, we seek to optimize the hyperparameter configuration for a specific task, on a single data set, in view of marrying multiple conflicting objectives. MTL, by contrast, seeks to optimize the HP configuration for multiple tasks, potentially using multiple datasets; while the performance metrics for the individual tasks can be seen as multiple simultaneous objectives, they are not necessarily  in conflict. 

Our work aims to provide an overview of the state-of-the-art in the field of \emph{multi-objective hyperparameter optimization} for machine learning algorithms, highlighting the approaches currently used in the literature, the typical performance measures used as objectives, and discussing remaining challenges in the field. To the best of our knowledge, our work presents the first comprehensive review of these multi-objective HPO approaches. Previous reviews  \citep{hutter2015beyond,luo2016review,yang2020hyperparameter,feurer2019hyperparameter,talbi2021automated}  mainly discuss single-objective HPO approaches, often focusing on particular contexts (such as biomedical data analysis), specific target algorithms (such as Deep Neural Networks) or specific approaches (Sequential Model-based Bayesian Optimization, multi-fidelity approaches). While two of the most recent surveys \citep{feurer2019hyperparameter,talbi2021automated} mention multi-objective HPO on the sidelines, they only list some examples or common strategies relevant to this topic, without discussing the actual approaches. 

The remainder of this article is organized as follows. Section \ref{sec:method} discusses the methodology used in the literature search. Section \ref{sec:problem_definition} formalizes the concepts of single- and multi-objective hyperparameter optimization and discusses the most commonly used performance measures in HPO algorithms. Section \ref{sec:algorithms} categorizes the existing methods for multi-objective hyperparameter optimization. Section \ref{sec:discussion_algorithms} discusses the pros and cons of the algorithms. Finally, section \ref{sec:conclusions} summarizes the findings, highlighting potential improvements and avenues for further research.

\section{Methodology}
\label{sec:method}

\begin{figure}[H]
\center
\includegraphics[width=11cm]{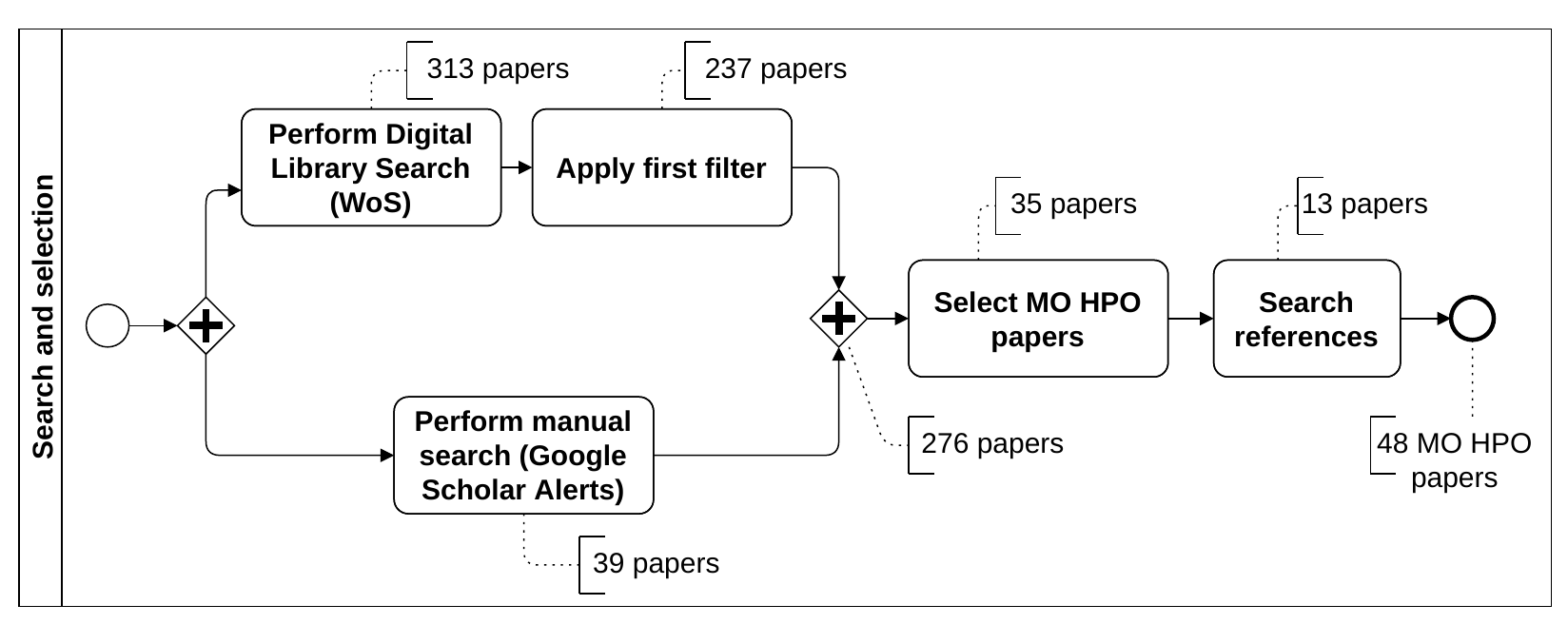} 
\caption{Overview of search and selection process}
\label{fig:search_diagram}
\end{figure}

Given the remarkable surge in publications on HPO since 2014, we focused on research published between 2014 and 2020. Figure \ref{fig:search_diagram} shows an overview of the search and selection process.

We first performed a WoS (Web of Science) search, using the search terms shown in Table \ref{tab:searchterms}. Although the main focus is on multi-objective HPO, we also consider the occurrence of the phrase ``single objective'' in the abstract (AB), as it is common to transform multiple objectives into a single objective by means of a scalarization function. As the use of surrogates is common in single-objective HPO for deep learning networks (e.g., \cite{wistuba2018scalable, sjoberg2019architecture, victoriaautomatic}), we also searched for articles mentioning the terms ``surrogate'', ``metamodel'', ``deep learning'', ``neural networks'', ``Gaussian process'', and ``kriging'' in the abstract. The choice of hyperparameters is also related to overfitting \citep{feurer2019hyperparameter}.  Finally, we also include the term ``constraint'', as the required performance targets (e.g., maximum memory consumption, training time \citep{stamoulis2018hyperpower, hu2019automatically}) may be presented as constraints in (multi-objective) HPO. We limited our search to publications (including conference proceedings, articles, book chapters, and meeting abstracts) in computer science-related categories (WC).

We subsequently completed the set of papers through (1) scanning suggestions of papers on Google Scholar alerts, and (2) a reference search. We limited the latter to electronic collections only, and solely considered journals/conference proceedings/workshop proceedings that were indexed on WoS (for the WoS journals, we included accepted preprints of forthcoming articles).

\begin{table}[!hbtp]
\center
\caption{Search term details}
\begin{tabular}{|l|l|}
\hline
\multicolumn{2}{|c|}{
\includegraphics[width=7.5cm]{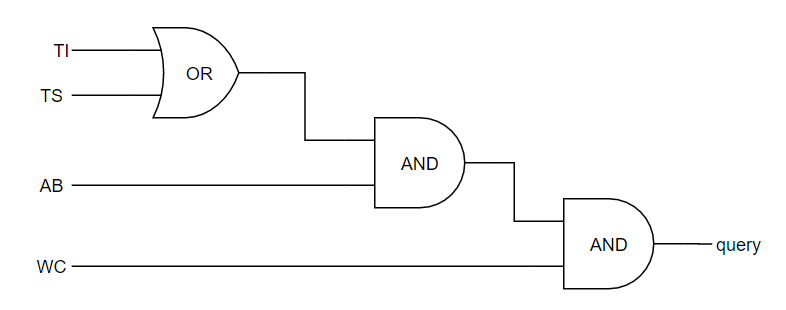} 
} \\ \hline
TI & \makecell[l]{hyperparameter optimization, hyperparameter tuning, parameter optimization, \\ hyperparameter, parameter tuning } \\ \hline
TS & hyperparameter optimization, hyperparameter tuning \\ \hline
AB & \makecell[l]{neural networks, deep learning, constraint, overfitting, multiobjective, \\multi objective, multi-objective, many-objective, many objective, single objective, \\surrogate, metamodel, gaussian process, kriging
} \\ \hline
WC & \makecell[l]{Computer Science, Artificial Intelligence \\ Computer Science, Information Systems \\ Computer Science, Theory \& Methods \\ Computer Science, Interdisciplinary Applications }\\ \hline
\end{tabular}
\footnotetext{TI: Title, TS: Topic, AB: Abstract, WC: Web of Science Categories}
\label{tab:searchterms}
\end{table}

The papers obtained through the WoS and manual search were manually filtered based on the title and abstract, to ensure they were related to the topic of discussion. We filtered out irrelevant papers, such as those that focus on the optimization of industrial processes \citep{chen2014effective}, meta-learning \citep{vanschoren2019meta}, optimization of internal parameters \citep{wawrzynski2017asd+}, and papers related to AutoML systems that are not focused on hyperparameter optimization (such as model selection algorithms \citep{van2015fast, silva2016hybrid} or pure feature selection methods \citep{hegde2020early}).
Neural Architecture Search (NAS) is usually considered a distinct category with its own methods and techniques for optimizing the structure of a neural network; hence, articles on NAS were only considered when the problem was addressed as an HPO problem. Articles focusing on more specific aspects of NAS (such as \cite{negrinho2019towards}) are beyond the scope of this research. 

\begin{figure}[H]
\center
\includegraphics[width=12cm]{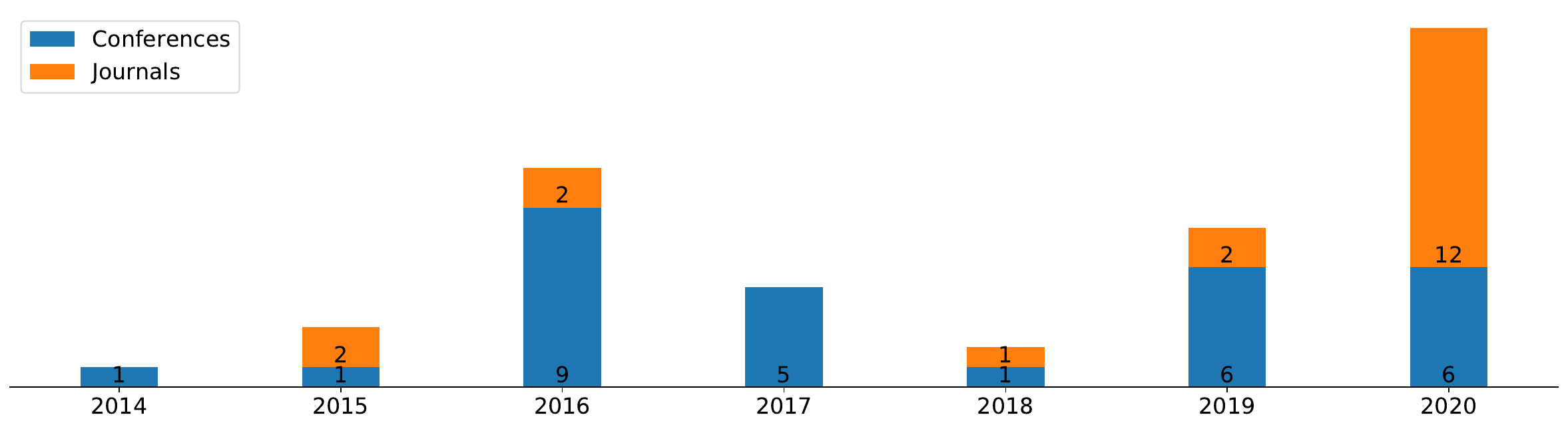} 
\caption{Number of articles that address multi-objective HPO, according to the publication source (2014-2020)}
\label{fig:history_MOHPO}
\end{figure}

A full read of the articles, combined with a reference search, resulted in a final selection of 48 relevant articles. Most of these articles (about 60\%) were published in conferences or workshops, though there has been an increase in scientific journal articles in 2020 (see Figure \ref{fig:history_MOHPO}); these were mainly published in Q1/Q2 journals belonging to the Computer Science field. 

\section{HPO: Concepts and performance measures}
\label{sec:problem_definition}

Section \ref{sec:concepts} provides an overview of the basic concepts related to HPO, while Section \ref{sec:obj} discusses the main performance measures (objectives) used in such optimization. Finally, Section \ref{sec:metrics} discusses the quality metrics used for comparing the performance of multi-objective HPO algorithms.

\subsection{HPO: Concepts and terminology}\label{sec:concepts}

\begin{figure}[!hbtp]

\includegraphics[width=12cm]{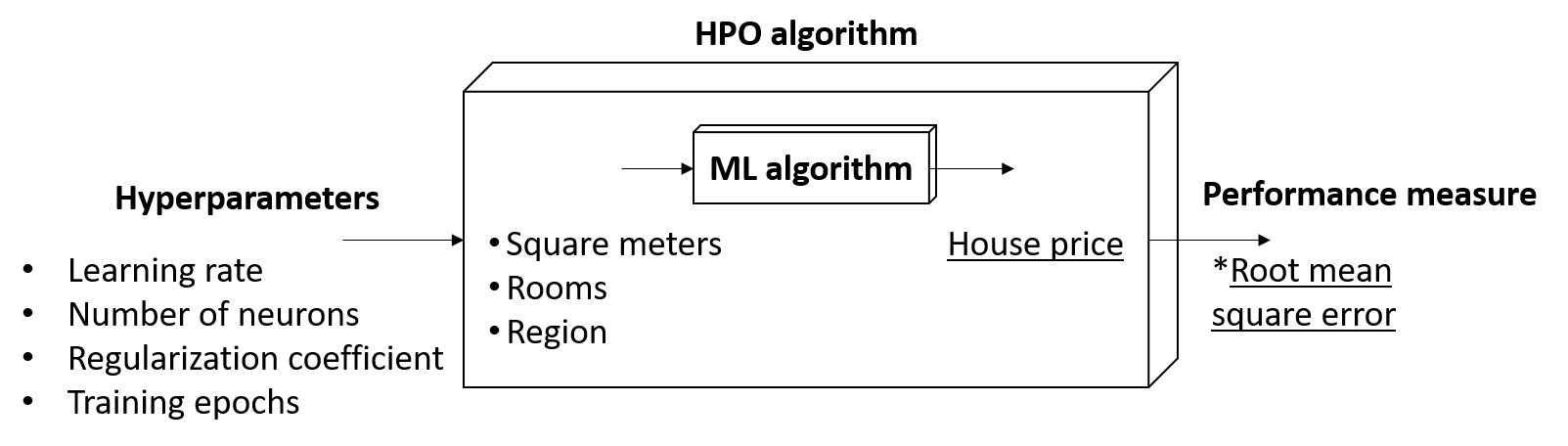}

\caption{Example of the interplay between the HPO algorithm and the target ML algorithm (in this case, an ANN for predicting house prices)}
\label{fig:hpo_algorithm}
\end{figure}

In mathematics and computer science, an algorithm is a finite sequence of well-defined instructions that, when fed with a set of initial inputs, eventually produces an output. 
Figure \ref{fig:hpo_algorithm} shows that in HPO, the optimization algorithm forms an ``outer'' shell of optimization instructions; the ``inner'' optimization refers to the training and cross-validation of the target ML algorithm (e.g., ANN, SVM, etc.). This inner optimization trains the target algorithm to perform the task it should perform (e.g., predicting house prices from a data set, using a set of features). In turn, the HPO algorithm takes the hyperparameters of the target ML algorithm as input and produces a number of performance measures as output (e.g., RMSE, energy consumption, etc.). The aim of the HPO algorithm is to optimize the set of hyperparameters, in view of obtaining the best possible outcomes for the performance measures considered.

More formally, the single-objective HPO problem can be formalized as follows. Consider a target ML algorithm $\mathcal{A}$ with $N$ hyperparameters, such that the \textit{n}-th hyperparameter has a domain denoted by $\Lambda_n$. The overall \textit{hyperparameter configuration space} is denoted as $\Lambda=\Lambda_1 \times \Lambda_2 \times ... \times \Lambda_N$. A vector of hyperparameters is denoted by $\boldsymbol{\lambda} \in \Lambda$, and an algorithm $\mathcal{A}$ with its hyperparameters set to $\boldsymbol{\lambda}$ is denoted by $\mathcal{A}_{\boldsymbol{\lambda}}$. In the case of HPO, the available data are split into a training set, a validation set, and a test set. The \emph{learning} process of the algorithm takes place on the training set ($\mathcal{D}_{train}$) and is validated on the validation set ($\mathcal{D}_{valid}$). We can then formalize the \emph{single-objective} HPO problem as \citep{feurer2019hyperparameter}: 

\begin{align}
    \min_{\boldsymbol{\lambda} \in \Lambda} V(\mathcal{L}\mid \mathcal{A}_{\boldsymbol{\lambda}}, \mathcal{D}_{train}, \mathcal{D}_{valid} )\nonumber
\end{align}

\noindent where $V(\mathcal{L} \mid \mathcal{A}_{\boldsymbol{\lambda}}, \mathcal{D}_{train}, \mathcal{D}_{valid} )$ is a validation protocol that uses a loss function $\mathcal{L}$ to estimate the performance of a model $\mathcal{A}_{\boldsymbol{\lambda}}$ trained on $\mathcal{D}_{train}$ and validated on $\mathcal{D}_{valid}$. Popular choices for the validation protocol $V(\cdot)$ are the holdout and cross-validation process (see \cite{bischl2012resampling} for an overview of validation protocols). Without loss of generality, we assume in the remainder of this article that the loss function should be minimized.

The previous definition can be readily extended to multi-objective optimization (see \cite{li2019quality}). Consider a multi-objective hyperparameter optimization problem with $N$ hyperparameters and a set $\mathbf{L}$ containing $m$ performance measures (objective functions). These can reflect the error-based performance of the algorithm, but also other metrics such as algorithm complexity (as detailed later in Section \ref{sec:obj}). 
The multi-objective HPO problem can then be formalized as follows (assuming that all performance measures should be minimized):

\begin{align}
\min_{\boldsymbol{\lambda} \in \Lambda} V(\mathbf{L} \mid \mathcal{A}_{\boldsymbol{\lambda}}, \mathcal{D}_{train}, \mathcal{D}_{valid} )  \nonumber
\end{align}


Typically, there is a trade-off among the different objectives: for instance, between the performance of a model and training time (increasing the accuracy of a model often requires larger amounts of data and, hence, a higher training time; see e.g., \cite{rajagopal2020deep}), or between different error-based measures (e.g., between confusion matrix-based measures \citep{tharwat2020classification} of a binary classification problem; see \cite{horn2016multi}). Considering these trade-offs is often crucial: e.g., in medical diagnostics \citep{de2002multi}, the simultaneous consideration of objectives such as sensitivity and specificity is essential to determine if the machine learning model can be used in practice. The goal in multi-objective HPO is to obtain the \emph{Pareto-optimal} solutions, i.e., those solutions for which none of the objectives can be improved without negatively affecting any other objective. In the decision space, the set of optimal solutions is referred to as the \emph{Pareto set}; in objective space, it yields the \emph{Pareto front} (or Pareto frontier). The Pareto-optimal solutions are also referred to as the \emph{non-dominated} solutions \citep{emmerich2018tutorial}. Ideally, these solutions should be \emph{diverse} (i.e., spread across the different areas of the Pareto front), while approximating this front as well as possible (i.e., showing \emph{convergence} to the Pareto front).

In (general) multi-objective optimization problems, the multiple objectives are often \textit{scalarized} into one single function, such that the problem can be solved as a single-objective problem. Care should be taken, though, when selecting the scalarization approach: e.g., not all approaches allow to detect non-convex parts of the front (see \cite{miettinen2002scalarizing} for further details about scalarization functions). Scalarization methods have also been applied in multi-objective HPO; see Section \ref{sec:algorithms} for further details.

\subsection{Multi-objective HPO: Typical objectives}
\label{sec:obj}

Table \ref{tab:objectives} and Table \ref{tab:objectives_non_error} show an overview and concise description of the performance measures occurring in the current literature on multi-objective HPO (Table \ref{tab:objectives} focuses on error-based measures, while Table \ref{tab:objectives_non_error} summarizes the non-error-based measures). These measures will reappear later in Section \ref{sec:algorithms}, when we categorize the different multi-objective HPO algorithms. As evident from Table \ref{tab:objectives}, for regression problems, the error-based metrics are commonly based on the squared errors; for classification problems, they are commonly related to the elements of the confusion matrix (True Positives (TP), False Negatives (FN), False Positives (FP), and True Negatives (TN)). 

Error-based measures are heavily used in multi-objective HPO, as they ensure a response from the model that is close to reality. Additionally, model complexity objectives are often included (following Occam's razor principle; \cite{blumer1987occam}), along with time-based metrics (e.g., training time on embedded devices) and/or (computational) cost objectives. The complexity of a neural network, for instance, is often estimated using the number of parameters (weights of the connections between neurons) \citep{liang2019evolutionary, lu2020nsganetv2, baldeon2020adaresu, calisto2020adaen}. The number of features can also be used as a complexity measure: see \citet{sopov2015self, martinez2017hybrid, binder2020multi, faris2020medical, bouraoui2018multi}. The more features the training algorithm has to consider, the more expensive it will be. On the other hand, considering fewer features may negatively affect the error-based performance of the algorithm. 

\begin{longtable}[H]{p{1.7cm}p{3cm}p{4.9cm}}
\caption{Error-based measures used in multi-objective HPO algorithms. The description given for the classification metrics assumes a two-class problem.} \label{tab:objectives} \\

\hline  
\multicolumn{1}{c}{\thead{\textbf{Type of}\\\textbf{problem}}} &
\multicolumn{1}{c}{\thead{\textbf{Performance}\\\textbf{measure}}} &
\multicolumn{1}{c}{\textbf{Description}} \\ \hline 
\endfirsthead

\hline  
\multicolumn{1}{c}{\textbf{Type of problem}} &
\multicolumn{1}{c}{\thead{\textbf{Performance}\\\textbf{measure}}} &
\multicolumn{1}{c}{\textbf{Description}}  \\ \hline  
\endhead

\hline \multicolumn{3}{r}{{Continued on next page}} \\ 
\endfoot
\endlastfoot

\multirow{7}{*}{Classification}  & Classification error  & \multicolumn{1}{c}{ $\frac{FN+FP}{P+N}$ }  \\ \cline{2-3} 

  & Recall / Sensitivity &  \multicolumn{1}{c}{ $\frac{TP}{TP+FN}$ }  \\ \cline{2-3}

& Precision &  \multicolumn{1}{c}{ $\frac{TP}{TP+FP}$ }  \\  \cline{2-3} 

 & Specificity & \multicolumn{1}{c}{ $\frac{TN}{TN+FP}$ }  \\ \cline{2-3} 
 
   & False positive rate (FPR) & \multicolumn{1}{c}{ $\frac{FP}{FP+TN}$ } \\ \cline{2-3} 
   
  & False negative rate (FNR) & \multicolumn{1}{c}{ $\frac{FN}{FN+TP}$ } \\ \hline
  & Reward of spiking trace & Custom measure defined for Spiking Neural Networks to measure how good was the network categorizing looming and non-looming stimuli at a given time \\ \hline
\multirow{3}{*}{Regression} & Root mean square error (RMSE) & \makecell{$\sqrt{\frac{\sum {\left ( Real - Prediction \right )}^{2}}{Total\:of\:observations}}$}  \\ \cline{2-3} 

 & Mean square error (MSE) & \makecell{$\frac{\sum {\left ( Real - Prediction \right )}^{2}}{Total\:of\:observations}$} \\ \cline{2-3} 
 
 & Sum square error (SSE) & \makecell{$\sum {\left ( Real - Prediction \right )}^{2}$}  \\ \hline
 
 Speech recognition &  Word error rate (WER) & Measures how different the recognized word is from the reference word  \\ \hline
 \multirow{2}{*}{\makecell[tl]{Image \\recognition}} & Segmentation accuracy & Computed as two times the area of overlap between two images divided by the total number of pixels in both images \\ \cline{2-3} 
  
  & Mutual Information & Measures the dependencies
between two images, or the amount of information that one image contains about the other \\ \hline 

\end{longtable}

\begin{longtable}[H]{p{3cm}p{8cm}}
\caption{Non error-based performance measures used in multi-objective HPO algorithms} \label{tab:objectives_non_error} \\

\hline  
\multicolumn{1}{c}{\textbf{Type}} &
\multicolumn{1}{c}{\textbf{Performance measure}}  \\ \hline  
\endfirsthead

\hline 
\multicolumn{1}{c}{\textbf{Type}} &
\multicolumn{1}{c}{\textbf{Performance measure}}  \\ \hline  
\endhead

\hline \multicolumn{2}{r}{{Continued on next page}} \\ 
\endfoot
\endlastfoot

\multirow{3}{*}{Complexity} & Number of floating points operations (FLOPs) or number of multiply-adds (MAdds) in NNs \\ \cline{2-2} 
 & Number of features used to train the ML algorithm \\ \cline{2-2} 
 & \multirow{2}{*}{Number of parameters (weights) in a NN} \\ \cline{1-1}
\multirow{5}{*}{Model size} &  \\ \cline{2-2} 
 & Number of neurons in NNs \\ \cline{2-2} 
 & Number of support vectors in SVM \\ \cline{2-2} 
 & The file size used to save a DNN \\ \cline{2-2} 
 & Number of models used in an ensemble \\ \hline
Time & Related to the target ML algorithm (Training and Prediction time, inference time on forward passes of ANN, decoding time), or the optimization \\ \hline
\multirow{2}{*}{\begin{tabular}[c]{@{}l@{}}Hardware-based \\measures \end{tabular}} & Memory footprint \\ \cline{2-2} 
 & Energy consumption \\ \hline
\multirow{2}{*}{Other} & Diversity measures in ensembles. \\ \cline{2-2} 
 & Outliers detected by a threshold-based algorithm.  \\ \hline

\end{longtable}


Metrics reflecting model size naturally depend on the target ML algorithm to be optimized (e.g., the number of neurons in a single-layer NN \citep{juang2014structure}, the number of support vectors in a SVM \citep{bouraoui2018multi}, the DNN file size \citep{shinozaki2020automated}, or the number of models used \citep{garrido2019predictive} for ensemble algorithms).
Alternatively, the number of floating point operations (FLOPs) in a NN can be used \citep{wang2019evolving,  wang2020particle,lu2020nsganetv2, chin2020pareco,loni2020deepmaker}. This metric is also used to reflect the energy consumption \citep{NIPS2015_ae0eb3ee}; likewise, the number of parameters in a NN is used as a measure for complexity as well as for model size. Both FLOPs and the number of parameters are sometimes used as memory consumption measures \citep{laskaridis2020hapi}, and can be combined with a time-based measure \citep{shah2016pareto}. Time-based measures can be related to the training phase \citep{tanaka2016automated, rajagopal2020deep, laskaridis2020hapi, lu2020nsganetv2}, the prediction phase \citep{hernandez2016predictive, abdolshah2019multi, garrido2019predictive}, the inference process on forwarding passes in ANNs \citep{kim2017nemo}, or the whole optimization process \citep{richter2016faster}. 

The increasing computational cost of Deep Learning models generally translates into higher hardware costs. As a result, optimization using both algorithm performance and hardware cost should be considered, especially for edge devices. Hardware-related costs can be measured in different ways; e.g., through energy consumption \citep{hernandez2016designing} or memory utilization \citep{chandrashekaran2016automated}. In many cases, these measures are estimated as a function of the hyperparameters. For instance, \cite{parsa2019pabo} present an abstract energy consumption model that depends on the neural network architecture (number of layers, number of outputs of each layer, kernel size, etc).

Some objectives encountered in the literature do not fall into any of the categories above. In Table \ref{tab:objectives_non_error}, they are grouped into the category ``Other'' (e.g., diversity measures for ensembles \citep{kuncheva2014combining}).


\subsection{Quality metrics for comparing multi-objective HPO algorithms}
\label{sec:metrics}

The surveyed literature presents different metrics to judge and/or compare the strengths and weaknesses of multi-objective HPO algorithms. 
The first set of quality metrics is related to the resulting Pareto front. Here, hypervolume is the most widely used \citep{horn2016multi, hernandez2016predictive, shah2016pareto, horn2017first, garrido2019predictive,lu2020nsganetv2}. It computes the volume of the area enclosed by the Pareto front and a reference point, specified by the user. \cite{binder2020multi} compute the \textit{generalization} dominated hypervolume, which is obtained by evaluating the non-dominated solutions of the validation set on the test set data. Other quality metrics based on the Pareto front are the difference in performance between each solution on the front and the single-objective version of the algorithm (holding the other objectives steady) \citep{chatelain2007multi}, the average distance (or Generational Distance) of the front to a reference set (such as the approximated true Pareto front obtained by exhaustive search, see  \cite{smithson2016neural}; or an aggregated front, see \cite{gulcu2021multi}), a coverage measure computed as the percentage of the solutions of an algorithm $A$ dominated by the solutions of another algorithm $B$ \citep{juang2014structure, li2004hybrid}, or metrics based on the shape of the Pareto front \citep{abdolshah2019multi} or its diversity  \citep{juang2014structure, li2004hybrid}. The latter can be computed using the spacing and the spread of the solutions: spacing evaluates the diversity of the Pareto points along a given front \citep{gulcu2021multi}, whereas spread evaluates the range of the objective function values (see \cite{zitzler2000comparison}).

Some authors use performance measures that do not relate to the quality of the front obtained; e.g., execution time \citep{parsa2019pabo, richter2016faster, horn2017first}, number of performance evaluations \citep{parsa2019pabo}, CPU utilization in parallel computer architectures \citep{richter2016faster}, measures that were not considered as an objective and that are evaluated in the Pareto solutions (usually, confusion matrix-based measures for classification problems; see \cite{salt2019parameter}), or measures that are specific for the HPO algorithm used (e.g., the number of new points suggested per batch is used by \cite{gupta2018exploiting} to evaluate the performance of the search executed during batch Bayesian optimization). 

\section{Multi-objective HPO algorithms: categorization}
\label{sec:algorithms}

In this section, we categorize the literature on multi-objective HPO algorithms based on the way in which the algorithms perform the search for the optimal solutions (i.e., the search methodology). We distinguish the following three categories: 
\begin{itemize}
    \item Metaheuristic-based optimization algorithms (Section \ref{sec:HPO_EA}): these algorithms use a metaheuristic to guide the search process, based on the empirically observed input/output observations.
    \item Metamodel-based optimization algorithms (Section \ref{sec:HPO_BO}): in these algorithms, a \emph{metamodel} is fit to the empirical input/output observations, and an acquisition function is used to search for the optimal HPO configurations.
    \item Hybrid algorithms (Section \ref{sec:HPO_Hybrid}): a metamodel is fit to the input/output observations, and a metaheuristic is used to guide the search for better solutions. 
\end{itemize}


\begin{figure}[H]
\center
\includegraphics[width=12cm]{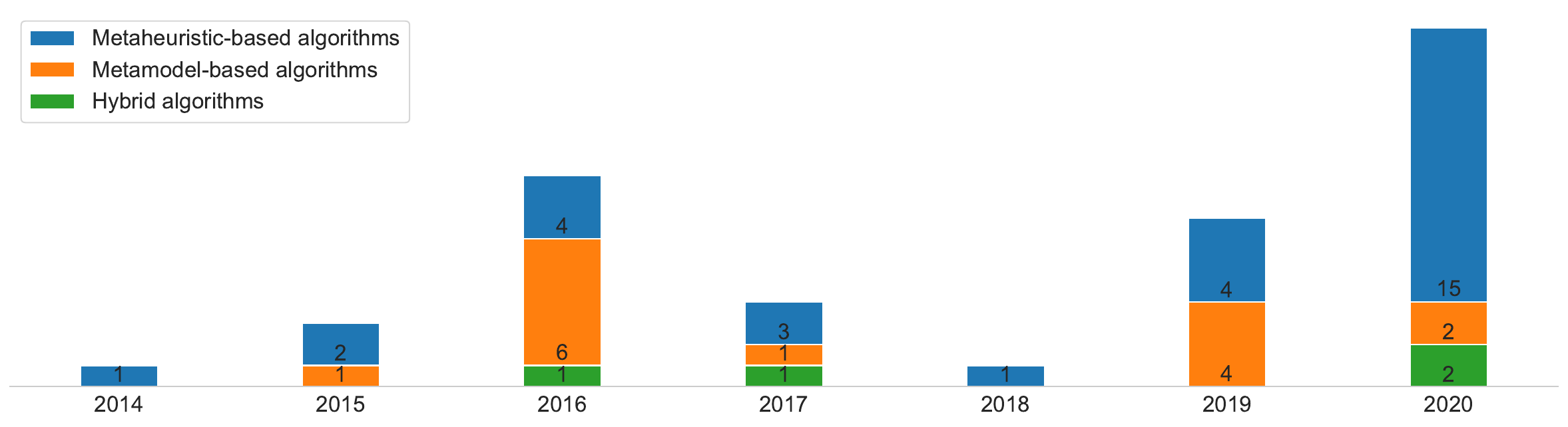} 
\caption{Multi-objective HPO algorithms: Number of articles per category  (2014-2020) }
\label{fig:type_per_year}
\end{figure}

\subsection{Metaheuristic-based HPO algorithms}
\label{sec:HPO_EA}

\begin{figure}[H]
\center
\includegraphics[width=14cm]{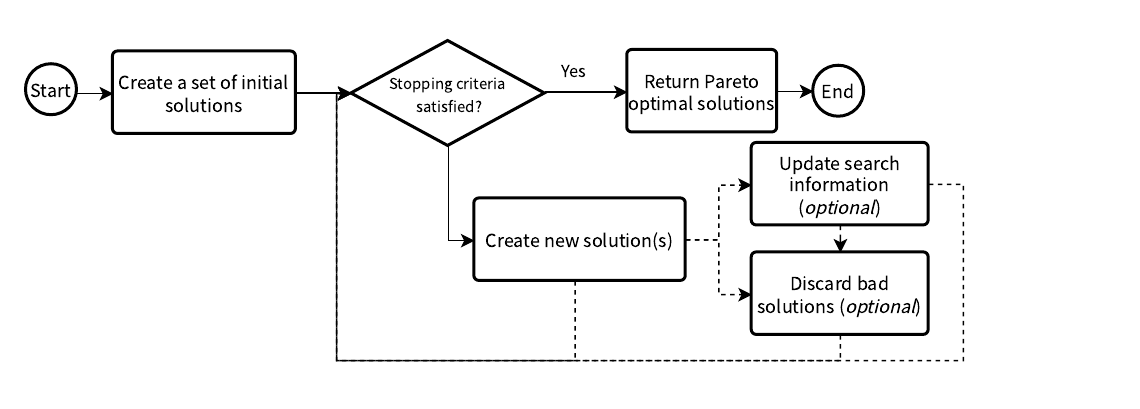} 
\caption{General procedure in metaheuristic-based MOO algorithms. }
\label{fig:metaheuristic}
\end{figure}

Heuristic search attempts to optimize a problem by improving the solution based on a given heuristic function or a cost measure \citep{russel2010}. A heuristic search method does not always guarantee to find the optimal solution but aims to find a good or acceptable solution within a reasonable amount of time and memory usage. Metaheuristics are algorithms that combine heuristics (which are often problem-specific) in a more general framework \citep{bianchi2009survey}. Figure \ref{fig:metaheuristic} summarizes the general procedure of a metaheuristic-based algorithm for multi-objective optimization (MOO). The algorithm generates new solution(s) starting from one or more initial solution(s). Depending on the algorithm, the information available from the search process so far (which can include updates in the sampling distribution used by the metaheuristic, \emph{or} other adjustments such as updates in the velocity vectors in Particle Swarm Optimization, or the pheromone paths in Ant Colony Optimization) can be updated before the next iteration starts, and/or bad solutions can be discarded. The process is repeated until a stop criterion is met.

While some metaheuristics start from a single initial solution  (e.g., Tabu Search \citep{glover1986future}), others (referred to as population-based algorithms) start from a set of solutions (e.g., Ant Colony Optimization \citep{dorigo2005ant} and Evolutionary Algorithms, e.g. Evolution Strategies and Genetic Algorithms \citep{mitchell1998introduction}). 

For ease of reference, Table \ref{tab:MO_EAHPO} gives an overview of the metaheuristic-based algorithms currently used in multi-objective HPO, while Table \ref{tab:MO_Comparison_meta} gives an overview of the experimental comparisons reported in these papers. Clearly, the most popular metaheuristic-based algorithm for multi-objective HPO is the Non-dominated Sorting Genetic Algorithm II (NSGA-II; \cite{deb2002fast}). This is not surprising, as genetic algorithms have shown to perform quite well in single-objective HPO settings: see, e.g., \cite{deighan2021genetic}, who showed that they cannot only obtain CNN configurations from scratch but can also refine state-of-the-art CNNs.
NSGA-II builds on the original NSGA algorithm \citep{srinivas1994muiltiobjective}; yet, it is computationally less expensive (a temporal complexity of $O(MN^2)$ versus $O(MN^3)$ for the original algorithm, where $M$ is the number of objectives and $N$ is the population size). Another important difference is the preservation of the best solutions, through an elitist selection according to the fitness and spread of solutions. \cite{ekbal2015joint} applied NSGA-II to jointly optimize hyperparameters and features, and demonstrated the superiority of the resulting models over others (trained with default hyperparameters, and using all the features included in a dataset). \cite{binder2020multi} observed analogous results optimizing a SVM, kkNN, and XGBoost. Yet, according to the generalization-dominated hypervolume, NSGA-II  performed slightly worse than ParEGO, a Bayesian optimization-based approach (see \cite{knowles2006parego} for further details). \cite{binder2020multi} thus suggest to prefer NSGA-II over ParEGO only when model evaluations are cheap and marginal degradation of performance is acceptable. 

Contrary to NSGA-II, the Multi-Objective Evolutionary Algorithm based on Decomposition (MOEA/D) \citep{zhang2007moea} uses \emph{scalarization} to solve the multi-objective HPO problem. Both MOEA/D and NSGA-II have shown to improve the accuracy of the resulting model compared with manual hyperparameter selection \citep{magda2017mooga, calisto2020adaen}. In \cite{baldeon2020adaresu}, MOEA/D is compared with a Bayesian Optimization approach (using Gaussian Process Regression with Expected Improvement as acquisition function), for tuning an adaptive convolutional neural network (AdaResU-Net) used for medical image segmentation. The use of MOEA/D resulted in a reduction in the number of parameters to train; the comparison is not really reliable, though, as the Bayesian approach was used in a single-objective optimizer, focusing only on segmentation accuracy and not on model size. The ENS-MOEA/D algorithm proposed by \cite{zhao2012decomposition} presents a further improvement to the original MOEA/D algorithm, by  adaptively adjusting the neighborhood size (as large neighborhood sizes favor more global search, while smaller sizes lead to more local search). \cite{zhang2020combined} apply this method to optimize the hyperparameters of a Variational Model Decomposition (VMD) procedure, used to pre-process time series for forecasting wind speeds. The authors prove that this yields better forecasts, yet they did not perform any comparison against other HPO procedures.

\begin{landscape}

\tiny
\begin{longtable}[H]{cp{1cm}p{0.5cm}p{1.3cm} p{1.5cm}p{1.5cm}p{1.5cm}p{1.5cm}p{1.5cm}p{1.5cm} p{1.5cm} }

\caption{Overview of Metaheuristic-based HPO algorithms. $N, D, C$ refer to the number of numeric, discrete, and categorical hyperparameters respectively. The use of scalarization is indicated in the first column (when relevant).}\label{tab:MO_EAHPO} \\
\hline  
\multirow{2}{*}{\thead{\textbf{HPO} \\\textbf{Algorithm}}} & 
\multirow{2}{*}{Ref.} & \multirow{2}{*}{HP} & 
\multirow{2}{*}{\thead{\textbf{Target ML}\\ \textbf{algorithm}} } & \multicolumn{6}{c}{\textbf{Performance measures}} & 
\multirow{2}{*}{\thead{\textbf{Application} \\ \textbf{field}}} \\ \cline{5-10}

 &  &  &  & \multicolumn{1}{c}{\textbf{Error}} & 
 \multicolumn{1}{c}{\textbf{Complexity}} & 
 \multicolumn{1}{c}{\textbf{Model size}} & 
 \multicolumn{1}{c}{\textbf{Time}} & 
 \multicolumn{1}{c}{\textbf{Hardware-based}} & 
 \multicolumn{1}{c}{\textbf{Other}} &  \\ \hline 
\endfirsthead

\hline  
\multirow{2}{*}{\thead{\textbf{HPO} \\\textbf{Algorithm}}} & 
\multirow{2}{*}{Ref.} & \multirow{2}{*}{HP} & 
\multirow{2}{*}{\thead{\textbf{Target ML}\\ \textbf{algorithm}} } & \multicolumn{6}{c}{\textbf{Performance measures}} & 
\multirow{2}{*}{\thead{\textbf{Application} \\ \textbf{field}}} \\ \cline{5-10}

 &  &  &  & \multicolumn{1}{c}{\textbf{Error}} & 
 \multicolumn{1}{c}{\textbf{Complexity}} & 
 \multicolumn{1}{c}{\textbf{Model size}} & 
 \multicolumn{1}{c}{\textbf{Time}} & 
 \multicolumn{1}{c}{\textbf{Hardware-based}} & 
 \multicolumn{1}{c}{\textbf{Other}} &  \\ \hline 
\endhead

\hline \multicolumn{11}{r}{{Continued on next page}} \\ 
\endfoot
\endlastfoot

\multirow{13}{*}{NSGA-II} & \cite{ekbal2015joint} & \makecell[tl]{\textbf{N:} 1, \\\textbf{D:} 1,\\ \textbf{C:} - } & \makecell[tl]{ CRF} & \multirow{2}{*}{ 
\makecell[l]{ Recall \\ Precision}} &\makecell[tc]{-}&\makecell[tc]{-}&\makecell[tc]{-}&\makecell[tc]{-}&\makecell[tc]{-}&  Mention recognition in texts \\ \cline{3-4}

& & \makecell[l]{\textbf{N:} 2, \\\textbf{D:} 2,\\ \textbf{C:} 1 }& \makecell[tl]{SVM} &  & \\ \cline{2-11}
 
& \cite{magda2017mooga} & \makecell[tl]{\textbf{N:} -, \\\textbf{D:} 2, \\\textbf{C:} - }&  Komi threshold algorithm  & \multirow{1}{*}{\begin{tabular}[c]{@{}l@{}} Mean error \end{tabular}} &\makecell[tc]{-}&\makecell[tc]{-}&\makecell[tc]{-}&\makecell[tc]{-}&\multirow{1}{*}{\begin{tabular}[c]{@{}l@{}} Number of \\outliers \end{tabular}}& Muscle onset detection \\ \cline{2-11}

& \cite{mostafa2020multi}  & \makecell[tl]{\textbf{N:} 3, \\\textbf{D:} 2, \\\textbf{C:} - }& CNN & \multirow{1}{*}{\begin{tabular}[c]{@{}l@{}} Accuracy \\ Recall \\Specificity \end{tabular}}  &\makecell[tc]{-}&\makecell[tc]{-}&\makecell[tc]{-}&\makecell[tc]{-}&\makecell[tc]{-}&
Apnea detection \\ \cline{2-11}
 
 & \cite{sopov2015self} & \makecell[tl]{\textbf{N:} -, \\\textbf{D:} 2, \\\textbf{C:} -} & MLP & \multirow{1}{*}{\begin{tabular}[c]{@{}l@{}}Classification \\rate \end{tabular}} &\makecell[tc]{-}&\multirow{1}{*}{\begin{tabular}[c]{@{}l@{}}Number of\\neurons \\ \end{tabular}} &\makecell[tc]{-}&\makecell[tc]{-}&\makecell[tc]{-}&
Emotion recognition \\  \cline{2-11}

& \cite{binder2020multi} & \makecell[tl]{\textbf{N:} -, \\\textbf{D:} 2, \\\textbf{C:} - }&  \multirow{1}{*}{SVM} & \makecell[tl]{Generalization \\error} & \makecell[tl]{Fraction of \\selected features} &\makecell[tc]{-}&\makecell[tc]{-}&\makecell[tc]{-}&\makecell[tc]{-}&
 \makecell[tl]{OpenML \\benchmark} \\\cline{3-4}

& & \makecell[tl]{\textbf{N:} 5, \\\textbf{D:} 4, \\\textbf{C:} - }& \makecell[tl]{XGBoost} & & & \\ \cline{3-4}

& & \makecell[tl]{\textbf{N:} -, \\\textbf{D:} 2, \\\textbf{C:} 1 }& \makecell[tl]{kkNN} &  &  &  &\\ \cline{2-11}

& \cite{shinozaki2020automated}  & \makecell[tl]{\textbf{N:} 7, \\\textbf{D:} 3, \\\textbf{C:} 1 }& Spoken Language Systems & \makecell[tl]{Word error \\ rate } &\makecell[tc]{-}&\makecell[tl]{DNN file size} &\makecell[tc]{-}&\makecell[tc]{-}&\makecell[tc]{-}&  Speech recognition \\ \cline{2-11}

& \cite{kim2017nemo} & \makecell[tl]{\textbf{N:} 2, \\\textbf{D:} 1, \\\textbf{C:} -  }& LeNet
 & \makecell[tl]{Accuracy } &\makecell[tc]{-}&\makecell[tc]{-}&\makecell[tl]{Inference time} &\makecell[tc]{-}&\makecell[tc]{-}&
Image recognition \\ \cline{2-11}
 
& \cite{bouraoui2018multi}  & \makecell[tl]{\textbf{N:} 4, \\\textbf{D:} 1, \\\textbf{C:} - }& \multirow{1}{*}{SVM} & \makecell[tl]{Accuracy} &\makecell[tl]{Number of \\features} & \makecell[tl]{Number of SV } &\makecell[tc]{-}&\makecell[tc]{-}&\makecell[tc]{-}&
 \multirow{1}{*}{UCI datasets} \\\cline{2-11}

& \cite{nabil2019deep}  & \makecell[tl]{\textbf{N:} 2, \\\textbf{D:} 2, \\\textbf{C:} 4 }& \makecell[tl]{GRU-based \\RNN} & \multirow{1}{*}{\begin{tabular}[ct]{@{}l@{}} Accuracy\\
FPR \end{tabular}} &\makecell[tc]{-}&\makecell[tc]{-}&\makecell[tc]{-}&\makecell[tc]{-}&\makecell[tc]{-}& 
 Electricity Theft Detection \\
\cline{2-11}

& \cite{loni2020deepmaker} & \makecell[tl]{\textbf{N:} -, \\\textbf{D:} 3, \\\textbf{C:} 3 } & CNN & \makecell[tl]{Accuracy } & \multicolumn{2}{l}{\makecell[tc]{Number of parameters }} &\makecell[tc]{-}&\makecell[tc]{-}&\makecell[tc]{-}& Image recognition \\ \cline{1-11}

\multirow{1}{*}{\begin{tabular}[ct]{@{}l@{}}
GA \\(scalarized \\objectives) \end{tabular}} & \cite{deighan2021genetic} & \makecell[tl]{\textbf{N:} 4, \\\textbf{D:} 6, \\\textbf{C:} - }& CNN & \makecell[tl]{Accuracy} &\multicolumn{2}{l}{\makecell[tc]{Number of parameters }} &\makecell[tc]{-}&\makecell[tc]{-}&\makecell[tc]{-}& Gravitational wave
classification \\ \cline{1-11}

\multirow{1}{*}{\begin{tabular}[ct]{@{}l@{}}
MOEA/D \\(scalarized \\objectives) \end{tabular}} & \cite{calisto2020adaen}  & \makecell[tl]{\textbf{N:} 1, \\\textbf{D:} 3, \\\textbf{C:} 3 }& AdaEn-net
 & \makecell[tl]{Segmentation \\ accuracy } &\multicolumn{2}{l}{\makecell[tc]{Number of parameters }} &\makecell[tc]{-}&\makecell[tc]{-}&\makecell[tc]{-}& Image segmentation \\ \cline{2-11}
 
& \cite{baldeon2020adaresu}  & \makecell[tl]{\textbf{N:} 2, \\\textbf{D:} 1, \\\textbf{C:} 2 }& AdaResU-net
 & \makecell[tl]{Segmentation \\ accuracy } &\multicolumn{2}{l}{\makecell[tc]{Number of parameters }} &\makecell[tc]{-}&\makecell[tc]{-}&\makecell[tc]{-}& Image segmentation \\ \cline{2-11} 
 
& \cite{zhang2016multiobjective}  & \makecell[tl]{\textbf{N:} 2, \\\textbf{D:} 1, \\\textbf{C:} - }& DBN ensembles
 & \makecell[lt]{Accuracy} &\makecell[tc]{-}&\makecell[tc]{-}&\makecell[tc]{-}&\makecell[tc]{-}&\makecell[tl]{Ensemble \\diversity }&
 Remaining useful life prediction \\  \cline{1-11}

\multirow{1}{*}{\begin{tabular}[ct]{@{}l@{}}
ENS-\\MOEA/D \end{tabular}} & \cite{zhang2020combined} & \makecell[tl]{\textbf{N:} 1, \\\textbf{D:} 2, \\\textbf{C:} - }& VMD & \multirow{1}{*}{\begin{tabular}[ct]{@{}l@{}} MSE\\
 Mutual \\Information \end{tabular}} &\makecell[tc]{-}&\makecell[tc]{-}&\makecell[tc]{-}&\makecell[tc]{-}&\makecell[tc]{-}& Wind speed prediction \\ \cline{1-11}
 
 \multirow{5}{*}{\begin{tabular}[ct]{@{}l@{}}CMA-ES \\ for MOO \end{tabular}} & \cite{tanaka2016automated}  & \makecell[tl]{\textbf{N:}19, \\\textbf{D:} 6, \\\textbf{C:} 2 }& NNLM & \makecell[tl]{Word error \\rate} &\makecell[tc]{-}&\makecell[tc]{-}&\makecell[tl]{Training time}&\makecell[tc]{-}&\makecell[tc]{-}&
 Speech recognition \\ \cline{2-11}
 
 & \cite{shinozaki2020automated}  & \makecell[tl]{\textbf{N:} 7, \\\textbf{D:} 3, \\\textbf{C:} 1 }& Spoken Language Systems & \makecell[tl]{Word error \\ rate} &\makecell[tc]{-}&\makecell[tl]{DNN file size}&\makecell[tc]{-}&\makecell[tc]{-}&\makecell[tc]{-}& Speech recognition \\\cline{2-11}

& \cite{qin2017evolution}  & \makecell[tl]{\textbf{N:} 6, \\\textbf{D:} 4, \\\textbf{C:} - }& \makecell[tl]{ NMT \\System}
 & \makecell[tl]{BLEU score} &\makecell[tc]{-}&\makecell[tc]{-}&\makecell[tl]{Validation time}&\makecell[tc]{-}&\makecell[tc]{-}&
Machine translation \\ \cline{2-11}
 
& \cite{ekbal2016simultaneous} & \makecell[tl]{\textbf{N:} -, \\\textbf{D:} 2, \\\textbf{C:} - }& CRF & \makecell[tl]{ Recall\\
Precision }  &\makecell[tc]{-}&\makecell[tc]{-}&\makecell[tc]{-}&\makecell[tc]{-}&\makecell[tc]{-}& Named entity recognition \\ \cline{3-4}
 
&  & \makecell[tl]{\textbf{N:} -, \\\textbf{D:} 1, \\\textbf{C:} - }& SVM &  &
  & \\ \cline{3-4}
  
&  & \makecell[tl]{\textbf{N:} -, \\\textbf{D:} 1, \\\textbf{C:} - }& MBL &  &
  & \\  \cline{1-11}
  
\multirow{2}{*}{OMOPSO}  & \cite{wang2019evolving, wang2020particle} & \makecell[tl]{\textbf{N:} -, \\\textbf{D:} 5, \\\textbf{C:} - }& Densenet-121 & \makecell[tl]{Accuracy} & FLOPs &\makecell[tc]{-}&\makecell[tc]{-}&\makecell[tc]{-}&\makecell[tc]{-}& Image classification \\  \cline{2-11}

& \cite{rajagopal2020deep} & \makecell[tl]{\textbf{N:} 3, \\\textbf{D:} 3, \\\textbf{C:} - }& CNN & \makecell[tl]{Accuracy} & FLOPs &\makecell[tc]{-}&\makecell[tc]{-}&\makecell[tc]{-}&\makecell[tc]{-}& Scene classification \\ \hline

\multirow{1}{*}{\begin{tabular}[ct]{@{}l@{}} PSO \\(scalarized \\objectives) \end{tabular}} & \cite{faris2020medical} & \makecell[tl]{\textbf{N:} 1, \\\textbf{D:} 2, \\\textbf{C:} - }& SVM & \makecell[tl]{ Error} & \makecell[tl]{Feature \\selection rate} &\makecell[tc]{-}&\makecell[tc]{-}&\makecell[tc]{-}&\makecell[tc]{-}& Medical specialties classification \\ \cline{1-11}

CoDeepNeat  & \cite{liang2019evolutionary} & \makecell[tl]{\textbf{N:} 2, \\\textbf{D:} 2, \\\textbf{C:} 3 }& CNN & \makecell[tl]{Error } & \multicolumn{2}{l}{\makecell[tc]{Number of parameters }}&\makecell[tc]{-}&\makecell[tc]{-}&\makecell[tc]{-}& Medical image classification \\ \cline{1-11}

\multirow{1}{*}{\begin{tabular}[ct]{@{}l@{}}
SPEA-II \\(scalarized \\objectives)\end{tabular}} & \cite{loni2019neuropower} & \makecell[tl]{\textbf{N:} -, \\\textbf{D:} 2, \\\textbf{C:} 4 }& CNN & \makecell[tl]{Accuracy} & \multicolumn{2}{l}{\makecell[tc]{Number of parameters }} &\makecell[tc]{-}&\makecell[tc]{-}&\makecell[tc]{-}& Image classification \\ \cline{1-11}

MADE  & \cite{pathak2020deep} & \makecell[tl]{\textbf{N:} 1, \\\textbf{D:} 5, \\\textbf{C:} 4 }& Bidirectional LSTM & \makecell[tl]{Recall\\
Specificity } &\makecell[tc]{-}&\makecell[tc]{-}&\makecell[tc]{-}&\makecell[tc]{-}&\makecell[tc]{-}& Classification \\  \cline{1-11}

\multirow{1}{*}{\begin{tabular}[ct]{@{}l@{}}
MODE \\(scalarized \\objectives)\end{tabular}}   & \cite{singh2020classification} & \makecell[tl]{\textbf{N:} 2, \\\textbf{D:} 5, \\\textbf{C:} 3 }& CNN & \makecell[tl]{Recall\\
Specificity } &\makecell[tc]{-}&\makecell[tc]{-}&\makecell[tc]{-}&\makecell[tc]{-}&\makecell[tc]{-}& Classification \\ \cline{1-11}

MO-RACACO & \cite{juang2014structure} & \makecell[tl]{\textbf{N:} -, \\\textbf{D:} 1, \\\textbf{C:} - }& Fuzzy Neural Networks & \makecell[tl]{RMSE} &\makecell[tc]{-}&\makecell[tl]{Number of \\neurons}&\makecell[tc]{-}&\makecell[tc]{-}&\makecell[tc]{-}& Regression \\ \cline{1-11}

MOSA & \cite{gulcu2021multi} & \makecell[tl]{\textbf{N:} -, \\\textbf{D:} 10, \\\textbf{C:} 4 }& CNN & \makecell[tl]{Accuracy} & FLOPs &\makecell[tc]{-}&\makecell[tc]{-}&\makecell[tc]{-}&\makecell[tc]{-}& Image classification \\ \cline{1-11}

\multirow{1}{*}{\begin{tabular}[ct]{@{}l@{}}
Nelder-Mead \\(scalarized \\objectives)\end{tabular}}  & \cite{albelwi2016automated} & \makecell[tl]{\textbf{N:} -, \\\textbf{D:} 7, \\\textbf{C:} - }& CNN & \makecell[tl]{Accuracy\\ Mutual \\information} &\makecell[tc]{-}&\makecell[tc]{-}&\makecell[tc]{-}&\makecell[tc]{-}&\makecell[tc]{-}& Image classification \\ \cline{1-11}

\end{longtable}
\normalsize
\end{landscape}

\tiny 
\begin{longtable}[H]{p{2cm}p{2.5cm}p{3cm}p{2.5cm}}

\caption{Experimental comparisons reported in the literature on metaheuristic-based HPO algorithms}\label{tab:MO_Comparison_meta} \\

\hline  
\multicolumn{1}{c}{\thead{\textbf{HPO} \\\textbf{Algorithm}}} &
\multicolumn{1}{c}{\textbf{Ref.}} &
\multicolumn{1}{c}{\thead{\textbf{Compared}\\ \textbf{against}}} &
\multicolumn{1}{c}{\thead{\textbf{Quality}\\ \textbf{metrics}}}\\\hline 
\endfirsthead

\hline  
\multicolumn{1}{c}{\thead{\textbf{HPO} \\\textbf{Algorithm}}} &
\multicolumn{1}{c}{\textbf{Ref.}} &
\multicolumn{1}{c}{\thead{\textbf{Compared}\\ \textbf{against}}} &
\multicolumn{1}{c}{\thead{\textbf{Quality}\\ \textbf{metrics}}}\\ \hline 
\endhead

\hline \multicolumn{4}{r}{{Continued on next page}} \\ 
\endfoot
\endlastfoot

\multirow{5}{*}{\makecell[tl]{NSGA-II}} & \cite{magda2017mooga} & Manual selection & Mean error (objective)\\ \cline{2-4}
 
 & \cite{sopov2015self} & SPEA \citep{zitzler1999multiobjective}, VEGA \citep{vega1985schaffer}, SelfCOMO-GA \citep{sopov2015self} & Classification rate and number of neurons (objectives) \\  \cline{2-4}

& \cite{binder2020multi} & ParEGO \citep{knowles2006parego} & Generalization error (objective) and hypervolume\\ \cline{2-4}

& \cite{shinozaki2020automated}  & CMA-ES \citep{hansen2003reducing} & WER and DNN file size (both objectives) \\ \cline{2-4}
 
& \cite{bouraoui2018multi}  & Grid search & Accuracy (objective) \\ \cline{1-4}

\makecell[lt]{
GA (scalarized \\objectives) } & \cite{deighan2021genetic} & GA variants & Scalarized fitness function \\ \cline{1-4}

\multirow{2}{*}{\makecell[lt]{
MOEA/D \\(scalarized \\objectives) }} & \cite{calisto2020adaen} & Manual selection & Segmentation accuracy and number of parameters (both objectives)\\ \cline{2-4}
 
& \cite{baldeon2020adaresu}  &  GP-EI \citep{snoek2012practical} & Performance measures \emph{not used as objectives} (Sensitivity, Mean surface distance, etc)\\ \cline{1-4} 

\makecell[lt]{CMA-ES for MOO } & \cite{shinozaki2020automated} & NSGA-II \citep{deb2002fast} & WER and DNN file size (both objectives) \\ \cline{1-4}

MO-RACACO & \cite{juang2014structure} & MO-EA \citep{juang2002tsk}, MO-ACOr \citep{socha2008ant} & Coverage metric and diversity in Pareto front\\ \cline{1-4}

\end{longtable}
\normalsize

The Covariance matrix adaptation-evolutionary strategy (CMA-ES) \citep{hansen2003reducing} is a population-based metaheuristic that differs from Genetic Algorithms in the use of a fixed-length real-valued vector as a gene (instead of the typical vector of binary components), and a multivariate Gaussian distribution to generate new solutions. Multi-objective CMA-ES can be formulated considering the dominance of solutions on the Pareto Frontier, to redefine the ranking function used to determine the best solution found so far (now a Pareto front) \citep{tanaka2016automated, qin2017evolution, shinozaki2020automated}.  \cite{shinozaki2020automated} optimize DNN-based Spoken Language Systems using this approach; the resulting networks had lower word error rates and were smaller than the networks designed by NSGA-II. Additionally, multi-objective CMA-ES generated smaller networks than the one obtained with single-objective CMA-ES (using the error-based measure as an objective to optimize). In our opinion, though, this last comparison does not make much sense, since network size did not appear as an objective in the single-objective setting. 

Analogous to Genetic Algorithms, Particle Swarm Optimization (PSO) \citep{eberhart1995new} works with a population of candidate solutions, known as \textit{particles}. Each particle is characterized by a velocity and a position. The particles search for the optimal solutions by continuously updating their position and velocity. Their movement is influenced not only by their own local best-known position but is also guided toward the best-known position found by other particles in the search space. A multi-objective PSO algorithm (OMOPSO) was developed by \cite{sierra2005improving}, using Pareto dominance and crowding distance to filter out the best particles. It employs different mutation operators which act on subsets of the swarm, and applies the $\epsilon$-dominance concept (see \cite{laumanns2002combining} for more details) to fix the size of the set of final solutions produced by the algorithm.

Strength Pareto Evolutionary Algorithm II (SPEA-II) \citep{zitzler2001spea2} adds several improvements to the original SPEA algorithm presented by \cite{zitzler1999multiobjective}. \cite{loni2019neuropower} used the algorithm to optimize six hyperparameters of a CNN, yielding more accurate and less complex networks than could be obtained with hand-crafted networks, or with NAS algorithms. 

Differential Evolution (DE) \citep{storn1997differential} is similar to Genetic Algorithms but differs in the way in which the solutions are coded (using real vectors instead of binary-coded ones) and, consequently, in the way in which the evolutionary operators are applied. Multi-Objective Differential Evolution (MODE) \citep{babu2007multi} selects the non-dominated solutions to generate new solutions on each iteration. To reduce the computational effort while maintaining accuracy, a memetic adaptive DE method (MADE) was developed by \cite{li2019parameter}. DE depends significantly on its control parameter settings. Therefore, MADE uses a historical memory of successful control parameter settings to guide the selection of future control parameter values \citep{tanabe2013success}. Additionally, a local search method (e.g., the Nelder-Mead simplex method (NMM) \citep{li2019parameter}, or chaotic local search \citep{pathak2020deep}) is employed to refine the solutions, and a ranking-based elimination strategy (using non-dominated and crowding distance sorting) is proposed to maintain the most promising solutions.

Ant Colony Optimization (ACO) \citep{dorigo1996ant} is inspired by the behavior of real ants; the basic idea is to model the HPO problem as the search for a minimum cost path in a graph. ACO algorithms can be applied to solve multi-objective problems, and may differ in three respects \citep{alaya2007ant}: (1) the way solutions are built, using only one \textit{pheromone structure} for an aggregation of several objectives, or associating a different pheromone structure with each objective \citep{iredi2001bi,gravel2002scheduling}; (2) the way in which solutions are updated \citep{iredi2001bi, baran2003multiobjective} and (3) the incorporation of existing problem-specific knowledge into the transition rule  that defines how to create new solutions from existing ones  \citep{gravel2002scheduling, doerner2004pareto}. The latter is included in a multi-objective version of ACO (MO-RACACO,  \cite{hsu2013multi}) for Fuzzy Neural Network (FNN) optimization \citep{juang2014structure}. The results showed that MO-RACACO outperformed other population-based MO algorithms (MO-EA,  \cite{juang2002tsk}; and MO-ACOr, \cite{socha2008ant}) in terms of the coverage measure obtained, yet it did not always obtain the best diversity values. 

Simulated annealing (SA) is a probabilistic technique for finding the global optimum of a single-objective problem \citep{kirkpatrick1983optimization}. \cite{gulcu2021multi} applied a multi-objective approach (MOSA) to optimize 14 hyperparameters of a CNN. The algorithm selects new solutions based on their relative  merit (measured by the dominance relationship) w.r.t. the current solutions.   

The Nelder-Mead simplex method (NMM) \citep{olsson1975nelder} has been applied by \cite{albelwi2016automated} to optimize seven hyperparameters for a CNN. As NMM is a single-objective optimization procedure, the objectives need to be scalarized (the authors used a weighted sum approach). NMM is a local optimization procedure, so it may get stuck in a local minimum. This may be avoided by running the algorithm from different starting points, which increases the probability of reaching the global minimum. Alternatively, modifications to the algorithm have been proposed (as in \cite{mckinnon1998convergence}) that allow the algorithm to escape from local minima, yet at the cost of a large number of iterations.

\subsection{Metamodel-based HPO algorithms}
\label{sec:HPO_BO}

\begin{figure}[H]
\center
\includegraphics[width=12cm]{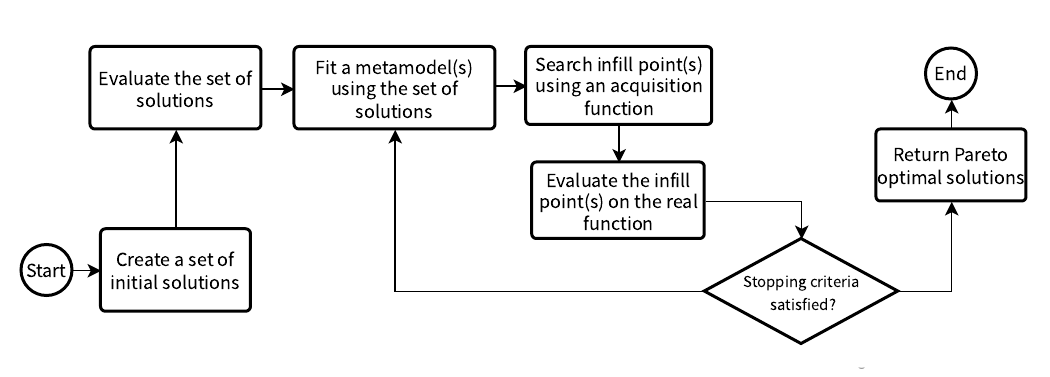} 
\caption{Generic optimization procedure in metamodel-based MOO algorithms. }
\label{fig:metamodel}
\end{figure}

Training a machine learning algorithm can be computationally expensive, e.g. due to the target algorithm's own structure (e.g., Deep Learning models), the amount and complexity of the data to process, resource limitations (execution time, memory and energy consumption, etc), and/or the type of training algorithm used. Therefore, different HPO approaches have been developed that employ less expensive models (referred to as metamodels or surrogate models) to emulate the computation of the real performance functions. The resulting algorithms have also been referred to as Efficient Global Optimization (EGO) or Bayesian Optimization (BO) algorithms, and use an \emph{acquisition function} or \emph{infill criterion} to guide the search. Figure \ref{fig:metamodel} summarizes the main steps in such an algorithm.  

The optimization starts with a set of initial points (input/output observations) to train the metamodel. Next, the acquisition function is used to select one or more new points (infill points) to be evaluated. The use of this acquisition function is a key element in the search (approaches that combine metamodels with metaheuristic search are referred to as \emph{hybrid} methods, and are discussed in Section \ref{sec:HPO_Hybrid}). The metamodel is updated with this new information (adding the new I/O observations to the initial set), and the procedure continues until a stopping criterion is met. 

For ease of reference, Table \ref{tab:MO_BOHPO} gives an overview of the metamodel-based algorithms currently used for multi-objective HPO, while Table \ref{tab:MO_comparison_metamodel} gives an overview of the experimental comparisons reported in this part of the literature. As evident from Table \ref{tab:MO_BOHPO}, most multi-objective HPO articles use a Gaussian Process (GP) metamodel. GPs use a covariance function, or kernel, to compute the \textit{spatial correlation} among several output observations for a given performance measure (i.e., a given objective of the HPO algorithm; see Figure \ref{fig:hpo_algorithm}). In this approach, it is assumed that HPO input configurations that differ only slightly from one another (i.e., they are \emph{close} to each other in the search space) are strongly positively correlated w.r.t. their outputs; as the configurations are further apart in the search space, the correlation dies out. The choice of the kernel in a GP is important, as it determines the shape of the assumed correlation function. In general, the most common kernels used in GP-based metamodels are the Gaussian kernel and the Matérn kernel \citep{ounpraseuth2008gaussian}. Using the kernel, the analyst can not only \emph{predict} the estimated outputs (i.e., in our case, the performance measures) at non-observed input locations (i.e., hyperparameter configurations), but can also estimate the \emph{uncertainty} on these output predictions. Both the predictions and their uncertainty are reflected in the acquisition function to search for new hyperparameter settings. We refer the reader to \cite{rojas2020survey} for a detailed review of acquisition functions, for (general, non-HPO related) single and multi-objective optimization problems.


\begin{landscape}
\tiny
\begin{longtable}[H]{cp{1cm}p{1cm}p{0.5cm}p{1.3cm}p{1.5cm}p{1.5cm}p{1.3cm}p{1cm}p{1.5cm}p{0.5cm}p{1.5cm} }
\caption{Overview of Metamodel-based HPO algorithms. $N, D, C$ refer to the number of numeric, discrete, and categorical hyperparameters respectively. The use of scalarization is indicated below the acquisition function (when relevant).}\label{tab:MO_BOHPO} \\

\hline 
\multirow{2}{*}{\thead{\textbf{Metamodel}}} & 
\multirow{2}{*}{\thead{\textbf{Acquisition}\\ \textbf{function}}} & 
\multirow{2}{*}{Ref.} & 
\multirow{2}{*}{HP} & 
\multirow{2}{*}{\thead{\textbf{Target ML}\\ \textbf{algorithm}} } & \multicolumn{6}{c}{\textbf{Performance measures}} & 
\multirow{2}{*}{\thead{\textbf{Application} \\ \textbf{field}}} \\ \cline{6-11}

& &  &  &  & \multicolumn{1}{c}{\textbf{Error}} & 
 \multicolumn{1}{c}{\textbf{Complexity}} & 
 \multicolumn{1}{c}{\textbf{Model size}} & 
 \multicolumn{1}{c}{\textbf{Time}} & 
 \multicolumn{1}{c}{\textbf{Hardware-based}} & 
 \multicolumn{1}{c}{\textbf{Other}} &  \\ \hline 
\endfirsthead

\hline
\multirow{2}{*}{\thead{\textbf{Metamodel}}} & 
\multirow{2}{*}{\thead{\textbf{Acquisition}\\ \textbf{function}}} & 
\multirow{2}{*}{Ref.} & 
\multirow{2}{*}{HP} & 
\multirow{2}{*}{\thead{\textbf{Target ML}\\ \textbf{algorithm}} } & \multicolumn{6}{c}{\textbf{Performance measures}} & 
\multirow{2}{*}{\thead{\textbf{Application} \\ \textbf{field}}} \\ \cline{6-11}

& &  &  &  & \multicolumn{1}{c}{\textbf{Error}} & 
 \multicolumn{1}{c}{\textbf{Complexity}} & 
 \multicolumn{1}{c}{\textbf{Model size}} & 
 \multicolumn{1}{c}{\textbf{Time}} & 
 \multicolumn{1}{c}{\textbf{Hardware-based}} & 
 \multicolumn{1}{c}{\textbf{Other}} &  \\ \hline  
\endhead

\hline \multicolumn{12}{r}{{Continued on next page}} \\ 
\endfoot
\endlastfoot

\multirow{10}{*}{\makecell[tl]{ Gaussian \\Process \\(deterministic \\observations)}} & EI (scalarized objectives)  & \cite{salt2019parameter} & \makecell[tl]{\textbf{N:}18, \\\textbf{D:} -, \\\textbf{C:} - }& SNN & \multirow{1}{*}{\begin{tabular}[ct]{@{}l@{}} Accuracy \\
Reward of the \\spiking trace\\
SSE \end{tabular}} &\makecell[tc]{-}&\makecell[tc]{-}&\makecell[tc]{-}&\makecell[tc]{-}&\makecell[tc]{-}&  Classification \\ \cline{2-12}

 &  \multirow{2}{*}{\makecell[tl]{Dominance \\rank of GP \\predictions}} & \cite{parsa2019pabo} & \makecell[tl]{\textbf{N:} 3, \\\textbf{D:} 6, \\\textbf{C:} - }& AlexNET & \multirow{2}{*}{\begin{tabular}[ct]{@{}l@{}} Error \end{tabular}}
 &\makecell[tc]{-}&\makecell[tc]{-}&\makecell[tc]{-}&\multirow{2}{*}{\begin{tabular}[ct]{@{}l@{}} Energy \\consumption \end{tabular}}&\makecell[tc]{-}&  Image classification \\ \cline{4-5}

& & & \makecell[tl]{\textbf{N:} -, \\\textbf{D:}11, \\\textbf{C:} - }& VGG19 &  &  & & \\ \cline{2-12}

& Preferences-based EHI & \cite{abdolshah2019multi} & \makecell[tl]{\textbf{N:} 4, \\\textbf{D:} 2, \\\textbf{C:} - }& NN & 
 \makecell[tl]{Prediction \\error }
 &\makecell[tc]{-}&\makecell[tc]{-}&\makecell[tl]{Prediction \\time }&\makecell[tc]{-}&\makecell[tc]{-}& Image classification \\ \cline{2-12}

& CEIPV & \cite{shah2016pareto} & \makecell[tl]{\textbf{N:} 1, \\\textbf{D:} 2, \\\textbf{C:} - }& NN & \makecell[tl]{Accuracy }
  &\makecell[tc]{-}&\makecell[tc]{-}&\makecell[tc]{-}&\makecell[tl]{Memory \\footprint }&\makecell[tc]{-}&  Classification \\ \cline{2-12}

& LCB & \cite{richter2016faster} & \makecell[tl]{\textbf{N:} -, \\\textbf{D:} 2, \\\textbf{C:} - }& SVM & \makecell[tl]{Classification \\error} &\makecell[tc]{-}&\makecell[tc]{-}&\makecell[tl]{Running \\time}&\makecell[tc]{-}&\makecell[tc]{-}&  Classification \\ \cline{2-12}

& UCB (scalarized objectives) & \cite{chin2020pareco} & \makecell[tl]{\textbf{N:} 6, \\\textbf{D:} -, \\\textbf{C:} - }& Slimmable NN & 
\makecell[tl]{Cross entropy \\loss}
&\makecell[tl]{FLOPs }&\makecell[tc]{-}&\makecell[tc]{-}&\makecell[tc]{-}&\makecell[tc]{-}&  Classification \\ \cline{2-12}
 
 & \multirow{2}{*}{PES} & \cite{hernandez2016predictive} & \makecell[tl]{\textbf{N:} 4, \\\textbf{D:} 2, \\\textbf{C:} - }& NN &\multirow{2}{*}{\begin{tabular}[ct]{@{}l@{}}Prediction \\error \end{tabular}}
 &\makecell[tc]{-}&\makecell[tc]{-}&\multirow{2}{*}{\begin{tabular}[ct]{@{}l@{}}Prediction \\time \end{tabular}}&\makecell[tc]{-}&\makecell[tc]{-}&  Image classification \\ \cline{3-12}
 
 & & \cite{garrido2019predictive} & \makecell[tl]{\textbf{N:} 2, \\\textbf{D:} 3, \\\textbf{C:} - }& Ensemble of Decision Trees & \makecell[tl]{Prediction \\error } &\makecell[tc]{-}&\makecell[tl]{Ensemble \\size }&\makecell[tc]{-}&\makecell[tc]{-}&\makecell[tc]{-}&  Classification  \\ \cline{3-12}

 & & \cite{hernandez2016designing} & \makecell[tl]{\textbf{N:} 4, \\\textbf{D:} 4, \\\textbf{C:} - }& NN & \makecell[tl]{Prediction \\error } &\makecell[tc]{-}&\makecell[tc]{-}&\makecell[tc]{-}&\makecell[tl]{Energy \\consumption}&\makecell[tc]{-}&  Classification \\ \cline{1-12}

\multirow{6}{*}{\makecell[tl]{Random \\Forest}} & \multirow{6}{*}{LCB} & \cite{binder2020multi} & \makecell[tl]{\textbf{N:} -, \\\textbf{D:} 2, \\\textbf{C:} - } & SVM & \multirow{3}{*}{\begin{tabular}[ct]{@{}l@{}} Generalization \\error\\ \end{tabular}}  &\multirow{3}{*}{\begin{tabular}[ct]{@{}l@{}} Fraction of\\ selected \\features \end{tabular}}&\makecell[tc]{-}&\makecell[tc]{-}&\makecell[tc]{-}&\makecell[tc]{-}& OpenML benchmark  \\ \cline{4-5}

 & & & \makecell[tl]{\textbf{N:} 5, \\\textbf{D:} 4, \\\textbf{C:} - }& XGBoost & & & & \\ \cline{4-5}
 
 & & & \makecell[tl]{\textbf{N:} -, \\\textbf{D:} 2, \\\textbf{C:} 1 }& kkNN & & & & \\ \cline{3-12}

 & & \cite{horn2016multi} & \makecell[tl]{\textbf{N:} -, \\\textbf{D:} 3, \\\textbf{C:} - } & SVM & \multirow{3}{*}{\begin{tabular}[ct]{@{}l@{}} FNR \\
FPR \end{tabular}}
  &\makecell[tc]{-}&\makecell[tc]{-}&\makecell[tc]{-}&\makecell[tc]{-}&\makecell[tc]{-}& OpenML Benchmark \\ \cline{4-5} 
 
& & & \makecell[tl]{\textbf{N:} -, \\\textbf{D:} 2, \\\textbf{C:} - }& Random Forest &  &  & & \\ \cline{4-5}

& & & \makecell[tl]{\textbf{N:} -, \\\textbf{D:} 2, \\\textbf{C:} - }& Logistic regression &  &  & &
\\ \cline{1-12}

\multirow{1}{*}{\begin{tabular}[ct]{@{}l@{}}
Tree \\Parzen \\Estimators \end{tabular}} & Modified EHV (scalarized objectives) & \cite{chandrashekaran2016automated} & \makecell[tl]{\textbf{N:} 3, \\\textbf{D:} 3, \\\textbf{C:} - }& NN-decoder & 
\makecell[tl]{Word error\\ rate}
 &\makecell[tc]{-}&\makecell[tc]{-}&\makecell[tl]{Decoding \\time}&\makecell[tl]{Memory \\consumption }&\makecell[tc]{-}&  Speech recognition \\ \cline{1-12}

\multirow{1}{*}{\makecell[tl]{ Gaussian \\Process \\(noisy  \\observations)}} & EHI & \cite{horn2017first} & \makecell[tl]{\textbf{N:} -, \\\textbf{D:} 3, \\\textbf{C:} - }& SVM & \multirow{1}{*}{\begin{tabular}[ct]{@{}l@{}} FNR \\
FPR \end{tabular}}  &\makecell[tc]{-}&\makecell[tc]{-}&\makecell[tc]{-}&\makecell[tc]{-}&\makecell[tc]{-}& OpenML benchmark \\ \cline{3-12}
 
 & & \cite{koch2015efficient} & \makecell[tl]{\textbf{N:} 2(5), \\\textbf{D:} -, \\\textbf{C:} - }& SVM & \makecell[tl]{Accuracy } &\makecell[tc]{-}&\makecell[tc]{-}& \makecell[tl]{Training \\time }&\makecell[tc]{-}&\makecell[tc]{-}& UCI benchmark \\ \cline{1-12}
 
\end{longtable}
\normalsize
\end{landscape}

\tiny 
\begin{longtable}[H]{p{2cm}p{2.5cm}p{3cm}p{2.5cm}}

\caption{Experimental comparisons reported in the literature on metamodel-based MO HPO algorithms}\label{tab:MO_comparison_metamodel} \\

\hline  
\multicolumn{1}{c}{\thead{\textbf{HPO} \\\textbf{Metamodel}}} &
\multicolumn{1}{c}{\textbf{Ref.}} &
\multicolumn{1}{c}{\thead{\textbf{Compared}\\ \textbf{against}}} &
\multicolumn{1}{c}{\thead{\textbf{Quality}\\ \textbf{metrics}}}\\\hline 
\endfirsthead

\hline  
\multicolumn{1}{c}{\thead{\textbf{HPO} \\\textbf{Metamodel}}} &
\multicolumn{1}{c}{\textbf{Ref.}} &
\multicolumn{1}{c}{\thead{\textbf{Compared}\\ \textbf{against}}} &
\multicolumn{1}{c}{\thead{\textbf{Quality}\\ \textbf{metrics}}}\\ \hline 
\endhead

\hline \multicolumn{4}{r}{{Continued on next page}} \\ 
\endfoot
\endlastfoot

\multirow{8}{*}{\makecell[tl]{ Gaussian \\process \\(deterministic\\ observations)}} & \cite{salt2019parameter} & Random search, DE \citep{storn1997differential}, SADE \citep{qin2008differential} & Performance measures not used as objectives \\ \cline{2-4}

 & \cite{abdolshah2019multi} & PESMO \citep{hernandez2016predictive}, SMS-EGO \citep{ponweiser2008multiobjective}, SUR \citep{picheny2014stepwise}, ParEGO \citep{knowles2006parego} & Descriptive analysis of the Pareto front \\ \cline{2-4}
 
 & \cite{parsa2019pabo} & Grid search, Random search, NSGA-II \citep{deb2002fast} & Execution time \\ \cline{2-4}
 
 & \cite{shah2016pareto} & ParEGO \citep{knowles2006parego}, Random search, GP-EHV & Hypervolume\\ \cline{2-4}
 
 & \cite{richter2016faster} & Random search & Classification error and running time (both objectives) and CPU usage \\ \cline{2-4}

 & \cite{hernandez2016predictive} & ParEGO \citep{knowles2006parego}, SMS-EGO \citep{ponweiser2008multiobjective}, SUR \citep{picheny2014stepwise} & Hypervolume \\ \cline{2-4}
 
  & \cite{garrido2019predictive} & Random search, BMOO \citep{feliot2017bayesian} & Hypervolume \\ \cline{2-4}
  
   & \cite{hernandez2016designing} & Random search, NSGA-II \citep{deb2002fast} & Hypervolume \\ \cline{1-4}

\makecell[tl]{ Gaussian \\process \\(noisy \\observations)} & \cite{horn2017first} & RTEA \citep{fieldsend2014rolling} , Random search & Hypervolume and runtime \\ \cline{2-4}

 & \cite{koch2015efficient} & SMS-EGO \citep{ponweiser2008multiobjective} , Latin Hypercube sampling & Hypervolume \\ \cline{1-4}
 
 \makecell[tl]{Random \\Forest } & \cite{horn2016multi} & SMS-EGO \citep{ponweiser2008multiobjective}, ParEGO \citep{knowles2006parego}, Random sampling, NSGA-II \citep{deb2002fast} & Hypervolume \\ \cline{1-4}
 
 TPE & \cite{chandrashekaran2016automated} & Random sampling, GP, Genetic Algorithm \citep{zames1981genetic} & WER, decoding time, and memory consumption (all ojectives) \\ \cline{1-4}
 
 
 
\end{longtable}
\normalsize

Table \ref{tab:MO_BOHPO} also shows the acquisition functions that have been used so far in multi-objective HPO. Clearly, the most popular one is Expected Improvement (EI, which was originally proposed by \cite{jones1998efficient}). The EI represents the expected improvement over the best outputs found so far, at an (arbitrary) non-observed input configuration. As EI was originally developed for single-objective problems, it is usually applied in multi-objective problems where the objectives are scalarized. \cite{salt2019parameter}, for instance, optimize a Spiking Neural Network (SNN) using a weighted function of three individual objectives (the accuracy, the sum square error of the membrane voltage signal, and the reward of the spiking trace). Three acquisition functions were studied; EI, Probability of Improvement (POI), and Upper Confidence Bound (UCB). The performance obtained with POI was significantly better than that obtained with EI and UCB, and overall, the BO-based approach required significantly fewer evaluations than evolutionary strategies such as SADE. 

Another way to use BO in multi-objective HPO is to fit a metamodel to each objective independently. \cite{parsa2019pabo} use such an approach in their Pseudo Agent-Based multi-objective Bayesian hyperparameter Optimization (PABO) algorithm; they use the dominance rank (based on the predictor values of each objective) as an infill criterion. This evidently yields different infill points for the respective objectives (in their case, an error-based objective and an energy-related objective). The infill point suggested for one objective function is then also evaluated for the other objective function, provided that it is not dominated by any previous HPO configuration analyzed. In this way, the algorithm speeds up the search for Pareto-optimal solutions. The experiments indeed demonstrated that PABO outperforms NSGA-II in terms of speed. 

Other authors have studied HPO problems when the performance measures are correlated \citep{shah2016pareto}, or when  one of the measures is clearly more important than the others \citep{abdolshah2019multi}. The algorithm proposed by \cite{shah2016pareto} models the correlations between accuracy, memory consumption, and training time of an ANN using a multi-output Gaussian process or Co-Kriging \citep{liu2018remarks}. The authors propose a modification to the expected hypervolume (EHV) that reflects these correlations; this modified EHV is then used as an acquisition function, preferring the infill point that increases the expected hypervolume of the Pareto front the most. The algorithm is compared to ParEGO, \citep{knowles2006parego}, random search, and a GP using the original EHV metric. The results suggest that the modified EHV criterion increases the speed of the optimization, requiring fewer iterations to converge to the Pareto optimal solutions. 

The MOBO-PC algorithm proposed by \cite{abdolshah2019multi} adjusts the \textit{Expected Hypervolume Improvement} (EHI) acquisition function to account for the probability that the novel HP configuration satisfies a set of user-defined preference-order constraints. In this way, it manages to focus its search on the Pareto solutions that are most relevant for the user, as opposed to the other algorithms that are used as a comparison in the paper (PESMO, \cite{hernandez2016predictive}; SMS-EGO, \cite{ponweiser2008multiobjective}; Stepwise Uncertainty Reduction, \cite{picheny2014stepwise}; and ParEGo, \cite{knowles2006parego}), which try to find solutions across the entire Pareto front. 

Other acquisition functions used in metamodel-based algorithms are the Lower Confidence Bound (LCB) or Upper Confidence Bound (UCB). These use a (user-defined) confidence bound to focus the search on local areas or explore the search space more globally. \cite{richter2016faster} use a multipoint LCB which simultaneously generates $q$ hyperparameter configurations. A GP is used to model the misclassification error and the logarithmic runtime. The results demonstrated an improvement in CPU utilization (and, thus, an increase in the number of hyperparameter evaluations)  within the same time budget. Confidence bounds are also used by \cite{chin2020pareco} to optimize the hyperparameters of Slimmable Neural Networks. The algorithm fits a GP to each individual performance measure, hence obtaining information to compute individual UCBs. These UCBs are then scalarized, and the resulting single objective function is minimized to obtain the next infill point. The proposed algorithm succeeds in reducing the complexity of the NNs studied; yet, the authors did not compare its performance with any other multi-objective HPO algorithms.

The Predictive Entropy Search (PES) criterion is used by multiple authors, as an infill criterion for different algorithms. \cite{hernandez2016predictive} use PESMO (multi-objective PES) to optimize a NN with six hyperparameters, in view of minimizing the prediction error and the training time. PESMO seeks to minimize the uncertainty in the \textit{location of the Pareto set}. The algorithm is compared with ParEGO, SMS-EGO, and SUR, showing that PESMO gives the best overall results in terms of hypervolume and the number of expensive evaluations required for training/testing the neural network. \cite{garrido2019predictive} use PESMOC (a modified version of PESMO which takes into account constraints)  to optimize an ensemble of Decision Trees. The experiments show that PESMOC is able to obtain better results than a state-of-the-art method for constrained multi-objective Bayesian optimization \citep{feliot2017bayesian}, in terms of the hypervolume obtained and the number of evaluations required. Finally, \cite{hernandez2016designing} used PES  to design a neural network with three layers. While most of the HPO methods collect data in a \textit{coupled} way by always evaluating all  performance measures jointly at a given input, these authors consider a \textit{decoupled} approach in which, at each iteration, the next infill configuration is selected according to the maximum value of the acquisition functions across all objectives. The results showed that this approach obtains better solutions (compared to NSGA-II and random search) when computational resources are limited; yet, the trade-offs found among the performance measures may be affected and one of the objectives can turn out to be prioritized over the others.

Random forests (RFs) \citep{ho1995random} are an ensemble learning method that trains a set of decision trees having low computational complexity. Each tree is trained with different samples, taken from the initial set of observations. For classification outputs, the RF uses a voting procedure to determine the decision class; for regression output, it returns the average value over the different trees. As for GP, RFs allows the analyst to obtain an uncertainty estimator for the prediction values. Some examples are the quantile regression forests method \citep{meinshausen2006quantile}, which estimates the prediction intervals, and the U-statistics approach \citep{mentch2016quantifying}. \cite{horn2016multi} use RFs as metamodel to optimize the hyperparameters of three ML algorithms: SVM, Random Forest, and Logistic regression. Using LCB as an acquisition function, the authors show that SMS-EGO and ParEGO outperform random sampling and NSGA-II. 

Whereas GP-based approaches model the density function of the resulting outcomes (performance measures) given a candidate input configuration, Tree-structured Parzen Estimators (TPE) \citep{bergstra2011algorithms} model the probability of obtaining an input configuration, given a condition on the outcomes. TPEs naturally handle not only continuous but also discrete and categorical inputs, which are difficult to handle with a GP. Moreover, TPE also works well for conditional search spaces (where the value of a given hyperparameter may depend on the value of another hyperparameter), and has demonstrated good performance on HPO problems for single-objective optimization \citep{bergstra2013making, thornton2013auto, falkner2018bohb}. While it can, in theory, also be applied to multi-objective settings by scalarizing the performance measures, \cite{chandrashekaran2016automated} obtained disappointing results when comparing this approach with random sampling, GP and Genetic Algorithms for optimizing an Augmented Tchebycheff scalarized function \citep{miettinen2012nonlinear} (using fixed weights) of three performance measures for ANNs: GP performed best, while TPE performed worst. Unfortunately, the authors reported the  performance based solely on the scalarized value of the three performance measures; they did not report on any other quality metrics, such as hypervolume. They also did not discuss the reason for the poor TPE performance, such that it remains unclear whether this is due to the scalarization function, or to the characteristics of the search space. A (non-scalarized) multi-objective version of TPE has been proposed by \cite{ozaki2020multiobjective} and is included in the software Optuna \citep{akiba2019optuna}.

Strikingly, the majority of current HPO algorithms routinely ignore the fact that the obtained performance measures are \emph{noisy}. The noise can be due to either the target ML algorithm itself (when it contains randomness in its procedure, such as a NN that randomly initializes the weights), but even if there is no randomness involved, there will be noise on the outcomes due to the use of \textit{k}-fold cross-validation during the training of the algorithm. This type of cross-validation is common in HPO: it involves the creation of different \emph{splits} of the data into a training and validation set. This process is repeated \textit{k} times; the performance measures of a given hyperparameter combination will thus differ for each split. Current HPO algorithms focus simply on the \emph{average} performance measures \emph{over the different splits} during the search for the Pareto-optimal points; the inherent uncertainty on these performance measures is ignored. \cite{horn2017first} are one of the few authors to highlight the presence of noise. The paper assumes, though, that noise is homogenous (i.e., it doesn't differ over the search space), and only focuses on different strategies for handling this noise. These strategies are used in combination with the SMS-EGO algorithm \citep{ponweiser2008multiobjective} and compared with the rolling tide evolutionary algorithm (RTEA) \citep{fieldsend2014rolling} and random search. The results show that simply ignoring the noise (by evaluating a given HPO combination only once, and considering the resulting performance measures as deterministic) performs poorly, even worse than a repeated random search. The best strategy is to reevaluate the (most promising) HP settings. According to the authors, this can likely be explained by the fact that the \emph{true} noise on the performance measures in HPO settings is heterogeneous (i.e., its magnitude differs over the search space). Reevaluation of already observed HP settings is then required to improve the reliability of the observed performance measures. The interested reader is referred to \cite{jalali2017comparison} for a discussion of the impact of noise magnitude and noise structure on the performance of (general) optimization algorithms. 

\cite{koch2015efficient} adapt SMS-EGO \citep{ponweiser2008multiobjective} and SExI-EGO
\citep{emmerich2011hypervolume} for noisy evaluations, to optimize the hyperparameters of a SVM. The authors again assume that the noise is homogenous, and compare the performance of both algorithms with different noise handling strategies (the reinterpolation method proposed by \cite{forrester2006design}, and static resampling). Both algorithms use the expected hypervolume improvement (EHI) as an infill criterion, though the actual calculation of the criterion is different (causing Sexi-EGO to require larger runtimes). The results show that both SMS-EGO and SExI-EGO work well with the reinterpolation method, yielding comparable results in terms of hypervolume.

\subsection{Hybrid HPO algorithms} 
\label{sec:HPO_Hybrid}

A limited number of papers have combined aspects of metamodel-based and population-based HPO approaches: these are referred to in Table \ref{tab:MO_HybridHPO}, summarizing their main characteristics. Table \ref{tab:MO_Comparison_hybrid} gives an overview of the experimental comparisons reported in these papers.

\cite{smithson2016neural} use an ANN as a metamodel to estimate the performance of the target ML algorithm. The neural network is embedded into a Design Space Exploration (DSE) metaheuristic, and is used to intelligently select new solutions that are likely to be Pareto optimal. The algorithm starts with a random solution, and iteratively generates new solutions that are evaluated with the ANN. DSE decides if the solution should be used to update the ANN knowledge, or should be discarded. Compared with manually designed networks from the literature, the proposed algorithm yields results with nearly identical performance, while reducing the associated costs (in terms of energy consumption).

The algorithm proposed by \cite{martinez2017hybrid} combines HPO with feature selection (as opposed to other algorithms, e.g., \cite{ekbal2015joint, leon2019convolutional, guo2019xgboost}). 
First, a GP (with UCB as an acquisition function) is used to obtain the best HPO setting (according to the RMSE), considering the full set of features. Next, a variant of GA (GA-PARSIMONY, \cite{sanz2015ga}) is used to select the best features of the problem, given the hyperparameters obtained in the first step. In this way, the final model has high accuracy and lower complexity (i.e., fewer features), and optimization time is significantly reduced. In our opinion, however, this approach is still suboptimal, as the two optimization problems (HPO and feature selection) are solved sequentially, instead of jointly. \cite{calisto2021emonas} use an evolutionary strategy combined with a Random Forest metamodel, to optimize 10 hyperparameters of a CNN. In the beginning of the optimization, the algorithm updates the population of solutions using the evolutionary strategy; only after some iterations, the selection of the new candidates is guided by the RF, which is updated each time with all new Pareto front solutions. The final networks found by the algorithm perform better than (or equivalent to) state-of-the-art architectures, while the size of the architectures and the search time are significantly reduced. 

Although most NAS algorithms are out of scope for this survey, we include the work by \cite{lu2020nsganetv2}, as it can be considered an HPO algorithm. The algorithm (NSGANetV2) simultaneously optimizes the architectural hyperparameters and the model weights of a CNN, using a bi-level approach consisting of NSGA-II combined with a metamodel. The metamodel is used to estimate performance measures, which are then optimized by an evolutionary algorithm (such approaches have also been applied successfully to non-HPO settings, see e.g.,  \cite{jin2011surrogate, dutta2020surrogate}). In the upper level of the optimization, the metamodel is built using an initial set of candidate solutions. In each iteration of the upper level, NSGA-II is executed on the metamodel to detect the Pareto-optimal HP settings (configuration of layers, channels, kernel size, and input resolution of the CNN). At the lower level, the weights of the CNN are trained on a subset of the Pareto-optimal solutions. The metamodel is then updated with the results of the actual performance evaluations. Four different metamodels were studied; Multilayer Perceptron (MLP), Classification and Regression Trees (CART), Radial Basis Functions (RBF), and GP. Given that none of them consistently outperformed the others, the authors propose to select the best metamodel in every iteration. On standard datasets (CIFAR-10, CIFAR-100, and ImageNet), the resulting algorithm matches the performance of state-of-the-art NAS algorithms \citep{lu2019nsga, mei2019atomnas}, but at a reduced search cost.

\begin{landscape}
\tiny
\begin{longtable}[H]{p{1.3cm}p{1cm}p{0.5cm}p{1.3cm} p{1.5cm}p{1.5cm}p{1.5cm}p{1.5cm}p{1.5cm}p{1.5cm} p{1.5cm} }
\caption{Overview of hybrid HPO algorithms. $N, D, C$ refer to the number of numeric, discrete, and categorical hyperparameters respectively. }\label{tab:MO_HybridHPO} \\

\hline 
\multirow{2}{*}{\thead{\textbf{HPO} \\\textbf{Algorithm}}} & 
\multirow{2}{*}{Ref.} & 
\multirow{2}{*}{HP} & 
\multirow{2}{*}{\thead{\textbf{Target ML}\\ \textbf{algorithm}} } & \multicolumn{6}{c}{\textbf{Performance measures}} & 
\multirow{2}{*}{\thead{\textbf{Application} \\ \textbf{field}}} \\ \cline{5-10}

 &  &  &  & \multicolumn{1}{c}{\textbf{Error}} & 
 \multicolumn{1}{c}{\textbf{Complexity}} & 
 \multicolumn{1}{c}{\textbf{Model size}} & 
 \multicolumn{1}{c}{\textbf{Time}} & 
 \multicolumn{1}{c}{\textbf{Hardware-based}} & 
 \multicolumn{1}{c}{\textbf{Other}} &  \\ \hline 
\endfirsthead

\hline 
\multirow{2}{*}{\thead{\textbf{HPO} \\\textbf{Algorithm}}} & 
\multirow{2}{*}{Ref.} & 
\multirow{2}{*}{HP} & 
\multirow{2}{*}{\thead{\textbf{Target ML}\\ \textbf{algorithm}} } & \multicolumn{6}{c}{\textbf{Performance measures}} & 
\multirow{2}{*}{\thead{\textbf{Application} \\ \textbf{field}}} \\ \cline{5-10}

 &  &  &  & \multicolumn{1}{c}{\textbf{Error}} & 
 \multicolumn{1}{c}{\textbf{Complexity}} & 
 \multicolumn{1}{c}{\textbf{Model size}} & 
 \multicolumn{1}{c}{\textbf{Time}} & 
 \multicolumn{1}{c}{\textbf{Hardware-based}} & 
 \multicolumn{1}{c}{\textbf{Other}} &  \\ \hline 
\endhead

\hline \multicolumn{7}{r}{{Continued on next page}} \\ 
\endfoot
\endlastfoot

ANN + DSE & \cite{smithson2016neural} & \makecell[tl]{\textbf{N:} 1, \\\textbf{D:} 2, \\\textbf{C:} 1 }& MLP & \multirow{2}{*}{\begin{tabular}[ct]{@{}l@{}}Accuracy \end{tabular}} &\makecell[tc]{Number of parameters} & &\makecell[tc]{-}&\makecell[tc]{-}&\makecell[tc]{-}&  Image classification \\ \cline{3-4}

 & & \makecell[tl]{\textbf{N:} 1, \\\textbf{D:} 4, \\\textbf{C:} 2 }& CNN & 
 &   & & \\ \cline{1-11}
 
 GP + GA Parsimony & \cite{martinez2017hybrid} & \makecell[tl]{\textbf{N:} 5, \\\textbf{D:} 3, \\\textbf{C:} - }& XGBoost & \makecell[tl]{RMSE } &\makecell[tl]{Number of \\features} &\makecell[tc]{-}&\makecell[tc]{-}&\makecell[tc]{-}&\makecell[tc]{-}& Image classification \\ \cline{1-11}

 \makecell[l]{(MLP/CART/\\RBF/GP) + \\NSGA-II} & \cite{lu2020nsganetv2} & \makecell[tl]{\textbf{N:} -, \\\textbf{D:} 4, \\\textbf{C:} - }& CNN & 
\makecell[tl]{Accuracy} &\makecell[tl]{MAdds } &\makecell[tc]{-}&\makecell[tc]{-}&\makecell[tc]{-}&\makecell[tc]{-}&  Image classification \\ \cline{1-11}

Random Forest + ES & \cite{calisto2021emonas} & \makecell[tl]{\textbf{N:} -, \\\textbf{D:} 6, \\\textbf{C:} 4 }& CNN & 
\makecell[tl]{Segmentation \\error} &\makecell[tc]{Number of parameters} & &\makecell[tc]{-}&\makecell[tc]{-}&\makecell[tc]{-}&  Image segmentation \\ \cline{1-11}

\end{longtable}
\normalsize
\end{landscape}

\tiny 
\begin{longtable}[H]{p{2cm}p{2.5cm}p{3cm}p{2.5cm}}

\caption{Experimental comparisons reported in the literature on hybrid MO HPO algorithms}\label{tab:MO_Comparison_hybrid} \\

\hline  
\multicolumn{1}{c}{\thead{\textbf{HPO} \\\textbf{Algorithm}}} &
\multicolumn{1}{c}{\textbf{Ref.}} &
\multicolumn{1}{c}{\thead{\textbf{Compared}\\ \textbf{against}}} &
\multicolumn{1}{c}{\thead{\textbf{Quality}\\ \textbf{metrics}}}\\\hline 
\endfirsthead

\hline  
\multicolumn{1}{c}{\thead{\textbf{HPO} \\\textbf{Algorithm}}} &
\multicolumn{1}{c}{\textbf{Ref.}} &
\multicolumn{1}{c}{\thead{\textbf{Compared}\\ \textbf{against}}} &
\multicolumn{1}{c}{\thead{\textbf{Quality}\\ \textbf{metrics}}}\\ \hline 
\endhead

\hline \multicolumn{4}{r}{{Continued on next page}} \\ 
\endfoot
\endlastfoot

ANN + DSE & \cite{smithson2016neural} & Exhaustive search & Generational Distance \\ \cline{1-4}
 
\makecell[tl]{(MLP/CART/RBF/\\GP) + NSGA-II} & \cite{lu2020nsganetv2} & NAS algorithms & Cumulative hypervolume, model size, and CPU/GPU latency \\ \cline{1-4}
 
\end{longtable}
\normalsize

\section{Multi-objective HPO algorithms: Pros and cons}
\label{sec:discussion_algorithms}

In this section, we discuss the weakness and strengths of the different algorithms. We focus on four different aspects: (1) the computational complexity of the algorithm, (2) the ability to accommodate high dimensional input spaces, (3) the ability to handle mixed input spaces, and (4) the ease of use of parallel computations. Unfortunately, none of the papers studied in this review provides explicit details on these aspects in the publication. In general, we often observed a surprising lack of detail with respect to many methodological aspects (such as the nature of the hyperparameters being optimized, the nature of the genetic operators and the design of the initial population in metaheuristic-based algorithms, the design of experiments used, the final Pareto-optimal solutions provided by the algorithm, etc.). In many cases, there is even no pseudocode provided for the algorithm, and detailed descriptions of novel metrics (if any) used to measure the performance of the target ML algorithm are lacking. This lack of detail is likely caused by the fact that most papers aim to solve a particular practical application and the hyperparameter optimization was usually not seen as the main contribution of the paper.

Consequently, the discussion in this section remains quite general, and relies largely on the results of our own independent research, based on the information found in \emph{methodological} papers for the algorithms considered. This information also allowed us to outline rough pseudocodes of the algorithms (which are presented in Appendix \ref{ap:pseudocode}). Although we emphasize (again) that these pseudocodes do \emph{not} necessarily reflect the accurate details of the algorithms, we find them helpful, in particular, to estimate the complexity of the algorithms. For black-box algorithms, this complexity can be measured by means of their \textit{worst-case expected running time} \citep{doerr2020complexity}. The running time (or \emph{optimization time} of an algorithm for a function $f$ is defined as the \emph{number of function evaluations} that the algorithm performs until (and including) the evaluation of an optimal solution for $f$. For HPO algorithms, the running time is largely proportional to the number of training and validation steps performed, as these are the most expensive steps in the HPO procedure. The training and validation steps need to be performed for \emph{each} HPO configuration studied by the HPO algorithm. Consequently, in what follows, we propose to use the (worst-case) number of HPO configurations evaluated by the algorithm as a proxy for the algorithm's expected worst-case running time. The result is expressed as a function $g(n,I,N)$, which is influenced by three parameters: (1) the number of initial HP configurations $n$ required to start the optimization (e.g., the size of the initial population in evolutionary algorithms, or the size of a Latin hypercube sample for Bayesian optimization), (2) the number of iterations $I$ allowed during the search, and (3) the number of new HPO configurations $N$ generated per iteration. Table \ref{tab:weakness_advantages} summarizes the results of our analysis.

Clearly, the number of costly function evaluations in a typical metamodel-based optimization is much lower than in a metaheuristic-based algorithm, as usually only a single new solution is evaluated in each iteration. MADE, the metaheuristic-based algorithm by \cite{pathak2020deep}, can be particularly expensive, as it performs a chaotic local search to generate $N$ additional solutions for each solution present in the population of a given iteration. However, using a metamodel to reduce the number of HP configurations that need to be evaluated does not ensure a lower execution time. For instance, the hybrid algorithm GP + GA\_Parsimony \citep{sanz2015ga} tries to optimize both hyperparameters and features used to train the ML model; the running time remains high, however, as the feature selection is performed in a separate phase after the HPO has been performed: this leads to a drastic increase in the number of HP configurations evaluated, compared with other algorithms such as NSGA-II and GP-based optimization. 

The use of parallel computations may be considered to decrease the total execution time of the optimization. For metaheuristic-based algorithms, this is usually implemented by parallelizing the evaluation of novel configurations in each population generation \citep{durillo2008study,wang2018particle}. Parallelization has been observed in metaheuristic-based optimization algorithms such as CMA-ES \citep{tanaka2016automated,qin2017evolution},  CoDeepNeat  \citep{liang2019evolutionary}, GA  \citep{deighan2021genetic}, and NSGA-II \citep{kim2017nemo}; it has also been suggested in \citep{albelwi2016automated,baldeon2020adaresu} for DNN optimization. Bayesian Optimization approaches, by contrast, are inherently serial as they use past observations to determine the next point(s) to sample. Parallelization can be used to some extent, though, e.g. in the evaluation of the initial set of configurations, or in batch BO \citep{richter2016faster,binder2020multi,horn2016multi}. Parallel computations can also be introduced during the training/validation of the ML algorithm (by training/validating  the model simultaneously on the different data splits in the cross-validation protocol \citep{mostafa2020multi}), or during the training of the metamodel (e.g., for Random Forests \citep{chen2016parallel} and for Gaussian Processes \citep{dai2014gaussian}).

The ability of an algorithm to handle mixed input spaces is not evident. For metaheuristic-based optimization procedures, for instance, this requires a proper coding of the solutions (e.g., the chromosomes in GAs or the particles in PSO), and consequently a reformulation of the evolutionary operators. For algorithms such as ACO, NMA, and CMA-ES, we expect that handling mixed search spaces is not straightforward, given that they were originally designed for a specific type of variables (ACO for discrete variables that can be easily structured in a graph, and NMA and CMA-ES for continuous variables). In metamodel-based optimization approaches using GPs, a proper kernel needs to be used to accommodate mixed input spaces. Metamodel-based approaches that rely on Random Forests or TPE, by contrast, can handle a mix of discrete, categorical, and numerical variables quite straightforwardly. 

To judge the ability of the algorithms to handle high dimensional search spaces, we relied on the findings of other studies (see the references in Table \ref{tab:weakness_advantages}). We categorize the results into poor (meaning that the ability to handle high dimensional search spaces is problematic), good, or unknown (meaning that no discussions on this aspect were found).

\begin{landscape}
\tiny
\begin{longtable}[H]{p{3cm}p{3.5cm}p{3cm}p{4cm}p{3cm} }
\caption{Analysis of pros and cons for the MO HPO algorithms studied. The number of HP configurations evaluated is expressed as a function $g(n, I, N)$, where $n$ is the number of initial solutions, $I$ is the number of iterations, and $N$ is the number of new HPO configurations per iteration. }\label{tab:weakness_advantages} \\

\hline 
\multicolumn{1}{c}{\thead{\textbf{HPO Algorithm}}} & 
\multicolumn{1}{c}{\thead{\textbf{Parallelization}}} & 
\multirow{1}{*}{\thead{\textbf{Number of HP}\\ \textbf{configurations evaluated}}} &
\multicolumn{1}{c}{\thead{\textbf{High dimensional input space}}} & 
\multicolumn{1}{c}{\thead{\textbf{Mixed search space}}} \\\\\\ \hline 
\endfirsthead

\hline 
\multicolumn{1}{c}{\thead{\textbf{HPO Algorithm}}} & 
\multicolumn{1}{c}{\thead{\textbf{Parallelization}}} & 
\multirow{1}{*}{\thead{\textbf{Number of HP}\\ \textbf{configurations evaluated}}} &
\multicolumn{1}{c}{\thead{\textbf{High dimensional input space}}} & 
\multicolumn{1}{c}{\thead{\textbf{Mixed search space}}} \\\\\\ \hline  
\endhead

\hline \multicolumn{5}{r}{{Continued on next page}} \\ 
\endfoot
\endlastfoot

GA, NSGA-II, CoDeepNeat, SPEA-II, ENS-MOEA,   MOEA/D, (MLP/CART/RBF/GP) + NSGA-II&  & \multirow{1}{*}{$g(n, I, N) = n+IN$} &  &  \\ \cline{1-1} \cline{3-3}

MADE &  & $g(n, I, N) = n + 2IN$ & \multirow{-4}{*}{Unknown} & \multirow{-4}{*}{\makecell[tl]{Requires implementation \\of specific genetic operators}} \\ \cline{1-1} \cline{3-5} 

OMOPSO,  PSO &  & $g(n, I, N) = n+IN$ & Poor  \citep{gad2022particle} & Requires a mixed-variable encoding scheme and specific reproduction methods \citep{wang2021particle} \\ \cline{1-1} \cline{3-5}

ACO &  & $g(n, I, N) = n+IN$ & Poor  \citep{ab2015comprehensive} & Mainly designed for discrete search spaces \\ \cline{1-1} \cline{3-5} 

CMA-ES for MOO & \multirow{-7}{*}{\makecell[tl]{Yes (for new \\solutions)}} & $g(n=0, I, N) = IN$ & Poor (due to the covariance matrix inversion \citep{shimizu2021cma}) & Requires adjustments  for non-real variables \\ \hline

GP-based metamodel & \multirow{2}{*}{\makecell[tl]{Yes (for initial set of solutions,\\ metamodel training) }} &  & Poor \citep{tripathy2016gaussian} & Requires specific kernel and specific optimization procedure for the acquisition function      \\ \cline{1-1} \cline{4-5} 

RF-based metamodel & &  & Good \citep{belgiu2016random} & No issues\\ \cline{1-2} \cline{4-5} 

TPE & Yes (for initial set of solutions) & \multirow{-3}{*}{$g(n, I, N=1) = n+I$} & Unknown & No issues 
\\ \hline

MODE & Yes (for each configuration, both in the initial set and new ones) & $g(n, I, N=1) = n+I$ & Unknown & Requires specific operators to generate new solutions \\ \hline

MOSA & Yes (to run the algorithm with different starting points, to explore new solutions in the neighborhood of the best solution)  & $g(n=0, I, N) = IN$ & Unknown & Requires specific operators for neighborhood generation\\ \hline

NMA & Yes (to run the algorithm with different starting points, \cite{joorabian2014optimal}) & $g(n, I, N) = n + 1 + I(N+1)$ & Unknown & Requires redefinition of the reflection, expansion, contraction and shrink operators  \\ \hline

GP + GA Parsimony & Yes (for each configuration, both in the initial set and new ones) & $g(n, I, N) = 2n+I(N+1)$ & Poor (due to the metamodel) & Requires specific kernel and specific optimization procedure for the acquisition function in the metamodel-based optimization phase  \\ \hline

ANN + DSE & Yes (for the initial set of solutions) & $g(n, I, N=1) = n+I$ & Unknown & Requires specific operator to generate new solutions \\ \hline

RF + ES & Yes (for the initial set of solutions, new solutions in each iteration, and metamodel training) & $g(n, I, N) = n+IN$ & Unknown & Requires specific operator to generate new solutions \\ \hline

\end{longtable}
\normalsize
\end{landscape}

\section{Conclusions and future research} 
\label{sec:conclusions}

This paper has reviewed the literature on multi-objective HPO algorithms, categorizing relevant papers into metaheuristic-based, metamodel-based, and hybrid approaches. The literature on MO HPO is not as abundant as on single-objective HPO; yet, MO HPO is highly relevant in practice. Taking a multi-objective perspective on HPO not only allows the analyst to optimize trade-offs between different performance measures, but it may also even yield \emph{better} solutions than the corresponding single-objective HPO problem. For instance, it has been shown that including complexity as an objective in multi-objective HPO does not necessarily compromise the loss-based performance of the ML algorithm w.r.t. the task for which it is trained: particularly, the minimization of the number of features used for training can \emph{improve} the performance of the ML algorithm \citep{sopov2015self, binder2020multi, bouraoui2018multi, faris2020medical}.


As the field of multi-objective HPO is gaining speed, it presents diverse opportunities for further research. We present recommendations here, distinguishing between (1) methodological recommendations (focusing on the use of more advanced optimization approaches), and (2) general recommendations (focusing on  shortcomings or pitfalls that currently occur in the literature, and that --in our opinion-- hamper the reproducibility, usability, and interpretability of the results). The recommendations are outlined in Table \ref{tab:opportunities}.

\begin{table}[!hbt]
\centering
\caption{Summary of research opportunities for Multi-Objective Hyperparameter Optimization} 
\begin{tabular}{p{3cm}p{8cm}}
\hline
\multicolumn{1}{c}{\makecell[c]{\textbf{Type} }} & \multicolumn{1}{c}{\textbf{Recommendations}} \\ \hline

Methodological & Use of hybrid algorithms\\
& Use of ensembles (of metamodels, acquisition functions, etc.)\\
& Multi-fidelity methods and/or bandit-based methods \\
& Use of early stopping criteria\\
& Use of algorithms that account for heterogeneous noise in performance objectives \\ \hline

General & Use of individual performance metrics instead of aggregated metrics \\
& Include a clear description of search space characteristics (type and range of considered HPs), algorithmic details (with pseudocode), performance objectives, and final optimal solutions obtained (optimal configurations, a quality metric for the Pareto front, etc.)\\
& Benchmark novel algorithms w.r.t. existing algorithms\\ 
\hline
\end{tabular}

\label{tab:opportunities} 
\end{table}

In the current literature, metaheuristic-based HPO approaches are clearly the most popular. This is quite striking, as such approaches require the evaluation of a large amount of HP configurations, and training/testing the target algorithm for any given HP configuration is usually the most expensive step in the HPO algorithm (due to, e.g., the $k$-fold cross-validation, the optimization steps required for the algorithm's internal parameters, the evaluation of potentially expensive performance measures such as energy consumption or inference time, etc.). Further research on hybrid HPO algorithms appears promising here. So far, research on these algorithms remains scarce; yet, one would expect that such algorithms combine the best of two worlds, providing low computational cost (as the metamodel provides inexpensive function evaluations) along with a heuristic search that avoids the challenge of optimizing an acquisition function. 

Current results have also demonstrated that using \emph{ensembles} of optimal HP configurations can yield improvements  \citep{ekbal2015joint, sopov2015self, ekbal2016simultaneous,zhang2016multiobjective}. Yet, this evidently increases the number of HP evaluations required. In future research, it may be promising to look at ensembles of multiple metamodels \citep{wistuba2018scalable, cho2020basic}, multiple acquisition functions \citep{cowen2020hebo}, or even multiple optimization procedures \citep{liu2020gpu}. 

Furthermore, multiple opportunities exist to extend recent advanced approaches for single-objective HPO towards multi-objective HPO. Recent research has shown potential benefits in studying cheaply available (yet lower fidelity) information, obtained for instance by evaluating only a fraction of the training data or a small number of iterations. Low fidelity methods such as bandit-based approaches \citep{li2017hyperband} have, to the best of our knowledge, not yet been applied in multi-objective HPO. Also, early stopping criteria \citep{dai2019bayesian} could be considered to ensure more intelligent use of the available computational budget. This has already been applied in single-objective optimization \citep{kohavi1995automatic,provost1999efficient}, by considering the algorithm’s learning curve: the training procedure for a given hyperparameter configuration is then stopped when adding further resources (training instance, iterations, training time, etc) is predicted to be futile. Early stopping criteria have also been used to reduce the overfitting level of the ML algorithm \citep{makarova2021overfitting}. To the best of our knowledge, none of these methodological approaches has been applied so far in multi-objective HPO algorithms.

Finally, apart from the work of \cite{koch2015efficient} and \cite{horn2017first}, the uncertainty in the performance measures is commonly ignored in HPO optimization. These two algorithms have mainly explored the impact of different noise handling strategies on the results of \emph{existing} algorithms, while it may be more beneficial to account for the noise by adjusting the metamodels used, and/or the algorithmic approach. Furthermore, they assume homogenous noise, which is likely not the case in practice. Stochastic algorithms (such as \citep{binois2019replication, gonzalez2020multiobjective}) can potentially be useful to determine the number of (extra)replications dynamically during HPO optimization, thus ensuring that computational budget is spent in (re-)evaluating the configuration that yields most information. 




Apart from these methodological recommendations, we also outline some general recommendations. To improve the interpretability of the results, we recommend using individual performance measures as objectives in HPO settings, rather than an aggregate measure such as the \textit{F-measure} (combining recall and precision for classification problems \citep{ekbal2015joint, ekbal2016simultaneous}) or the \textit{Area Under the Curve} measure (AUC), which combines the False Positive rate and the True Positive rate. Such aggregated measures reflect a fixed relationship between the individual measures, which may result in solutions that perform really well on the aggregated measure (for instance, the F-measure), but are suboptimal for the individual measures (recall and precision). Moreover, the aggregation of multiple performance measures into a single objective by means of scalarization should be done carefully, as not all scalarization methods (e.g., weighted sum) allow the detection of all parts of the Pareto front. The Augmented Tchebycheff function \citep{miettinen2012nonlinear}, for instance, is recommended when the front contains non-convex areas. The nonlinear term in the scalarization function ensures that these areas can be detected, while the linear term ensures that weak Pareto optimal solutions are less rewarded (see \cite{miettinen2002scalarizing} for a further discussion on scalarization functions). 

Furthermore, we noticed a surprising lack of detail in the current HPO papers (i.e., in the description of the methodological approaches, the experimental designs, and the corresponding results). To improve the reproducibility of the research, and facilitate comparisons among different HPO algorithms, we recommend a clear description and analysis of four basic elements in each future HPO research paper: (1) search space characteristics (type and range of the considered HPs), (2) algorithmic details (accompanied by pseudocode), (3) description/definition of performance objectives, (4) details on the final optimal solutions obtained for the test problems (optimal HPO configurations, quality metrics for the Pareto front, etc.).  

Finally, we noticed that only about half of the papers studied benchmark the algorithm under study w.r.t. other existing algorithms. Such experimental comparisons have substantial added value for the research community. We therefore clearly advocate their inclusion in future multi-objective HPO research.

\backmatter

\bmhead*{Declarations of interests}
The authors declare that they have no known competing financial interests or personal relationships that could have influenced the work reported in this paper.

\bmhead*{Acknowledgments}
 This work was supported by the Flanders Artificial Intelligence Research Program (FLAIR), and by the Research Foundation Flanders (FWO Grant 1216021N). The authors would like to thank Gonzalo Nápoles from Tilburg University for his comments on a previous version of this paper.
 
\begin{appendices}

\section{Pseudocodes of MO-HPO algorithms}\label{secA1}
\label{ap:pseudocode}

The details of the optimization algorithms are provided as a pseudocode. This was obtained from the description included in the papers surveyed and the paper where the algorithm was proposed initially. The function indicating the number of HP configurations evaluated during the optimization is included as part of the heading of the pseudocode.

\begin{megaalgorithm}
\caption{Metamodel-based optimization $g(n, I, N=1) = n+I$ }\label{alg:BO_opt}
\begin{algorithmic}[1]
\Require $n:$ initial design, $I$: number of iterations

\State $P \leftarrow \{\}$ 
\For{$i=1$ \TO $n$} \Comment{\textbf{Sample initial HP configurations}}
    \State $hp \leftarrow$ Generate HP configuration
    \State $f_{hp} \leftarrow$ Evaluate performance of $hp$
    \State $P \leftarrow P \cup \{hp, f_{hp}\}$
\EndFor
\For{$i=1$ \TO $I$}
    \State $M \leftarrow$ Train metamodel using $P$
    \State $new_{hp} \leftarrow$ Obtain a new HP configuration by optimizing an acquisition function using the metamodel predictions

    \State $f_{hp} \leftarrow$ Evaluate performance of $new_{hp}$
    \State $P \leftarrow P \cup \{new_{hp}, f_{hp}\}$
\EndFor
    
\Return HP configurations in the Pareto front 
\end{algorithmic}
\end{megaalgorithm}  

\begin{megaalgorithm}
\caption{NSGA-II, OMOPSO, PSO, SPEA-II, MO-RACACO, CoDeepNEAT, ENS-MOEA/D, MOEA/D, GA with scalarized objectives, and the hybrid algorithm (MLP/CART/RBF/GP) + NSGA-II $g(n, I, N) = n+IN$ }\label{alg:NSGA_II}
\begin{algorithmic}[1]
\Require $n:$ population size, $I$: number of iterations, $N$: number of new configurations per iteration 

\State $P \leftarrow \{\}$ 
\For{$i = 1$ \TO $n$} \Comment{\textbf{Create initial population}}
    \State $hp \leftarrow$ Generate HP configuration
    \State $f_{hp} \leftarrow$ Evaluate performance of $hp$
    \State $P \leftarrow P \cup \{hp, f_{hp}\}$
\EndFor

\For{$i=1$ \TO $I$}
    \For{$q=1$ \TO $N$} \Comment{\textbf{Generate new HP configurations}}
    
        \State $new_{hp} \leftarrow$ Generate HP configuration 
        \State $f_{new_{hp}} \leftarrow$ Evaluate performance of $new_{hp}$
        \State $P \leftarrow P \cup \{new_{hp}, f_{new_{hp}}\}$
    \EndFor
    
    \If{NSGA-II or CoDeepNeat or GA or hybrid algorithm} 
        \State $P \leftarrow$ Select HP configurations considering non-dominated and crowding distance sorting
    \EndIf
     \If{MOEA/D or ENS-MOEA/D} 
        \State $P \leftarrow$ Select HP configurations considering non-dominated sorting
    \EndIf
    \If{OMOPSO or PSO} 
        \State Update velocity and particle position 
    \EndIf
    \If{MO-RACACO} 
        \State Update pheromone paths 
    \EndIf
    
\EndFor

\Return HP configurations in the Pareto front 
\end{algorithmic}
\end{megaalgorithm}

\begin{megaalgorithm}
\caption{Random Forest + ES $g(n, I, N) = n + IN$ }\label{alg:rf_es}
\begin{algorithmic}[1]
\Require $n:$ initial solutions, $I$: number of iterations, $N$: number of new configurations per iteration 

\State $P \leftarrow \{\}$ 
\State $STP \leftarrow \{\}$ \Comment{\textbf{Set of observed HP configurations to train the metamodel}}

\For{$i=1$ \TO $n$} \Comment{\textbf{Create initial Population}}
    \State $hp \leftarrow$ Generate an HP configuration
    \State $f_{hp} \leftarrow$ Evaluate performance of $hp$
    \State $P \leftarrow P \cup \{hp, f_{hp}\}$
\EndFor
\State $STP \leftarrow STP \cup P$

\For{$i=1$ \TO $\frac{I}{2}$} \Comment{\textbf{Perform Evolutionary Strategy without the metamodel during the first half of the iterations}}
    \For{$q=1$ \TO $N$} 
        \State $new_{hp} \leftarrow$ Generate new HP configuration by applying genetic operators
        \State $f_{new_{hp}} \leftarrow$ Evaluate performance of $new_{hp}$
        \State $P \leftarrow P \cup \{hp, f_{new_{hp}}\}$
        \State $STP \leftarrow STP \cup \{hp, f_{new_{hp}}\}$
    \EndFor
    
    \State $P \leftarrow$ Select HP configurations considering non-dominated and crowding distance sorting
\EndFor

\For{$i=1$ \TO $\frac{I}{2}$} \Comment{\textbf{Perform Evolutionary Strategy using the metamodel during the second half of the iterations}}
    \For{$q=1$ \TO $N$} 
        \State Train a metamodel (Random Forest) using $STP$
    
        \State $hp \leftarrow$ Generate a HP configuration using genetic operators and the metamodel prediction
        \State $f_{hp} \leftarrow$ Evaluate performance of $hp$
        \State $P \leftarrow P \cup \{hp, f_{hp}\}$
        \State $STP \leftarrow STP \cup \{hp, f_{hp}\}$
    \EndFor
    
    \State $P \leftarrow$ Select HP configurations considering non-dominated and crowding distance sorting
\EndFor

\Return HP configurations in the Pareto front 
\end{algorithmic}
\end{megaalgorithm}

\begin{megaalgorithm}
\caption{MODE for scalarized objectives $g(n, I, N=1) = n+I$ }\label{alg:mode}
\begin{algorithmic}[1]
\Require $n:$ initial design, $I$: number of iterations

\State $P \leftarrow \{\}$ 
\For{$i=1$ \TO $n$} \Comment{\textbf{Sample initial HP configurations}}
    \State $hp \leftarrow$ Generate HP configuration
    \State $f_{hp} \leftarrow$ Evaluate performance of $hp$
    \State $P \leftarrow P \cup \{hp, f_{hp}\}$
\EndFor
\For{$i=1$ \TO $I$}

    \State $challenger \leftarrow$ Select a random configuration from $P$
    \State $a, b, c \leftarrow$ Select three random configurations from $P$
    
    \State $new_{hp} \leftarrow$ Obtain a new HP configuration as a linear combination of $a, b, c$

    \State $f_{new_{hp}} \leftarrow$ Evaluate performance of $new_{hp}$
    \If{$f_{new_{hp}}$ is better than $f_{challenger}$} 
        \State replace $challenger$ with $new_{hp}$ in $P$
    \EndIf
\EndFor
    
\Return HP configurations in the Pareto front
\end{algorithmic}
\end{megaalgorithm}  

\begin{megaalgorithm}
\caption{ANN + DSE $g(n, I, N=1) = n + I$ }\label{alg:ann_dse}
\begin{algorithmic}[1]
\Require $n:$ initial design, $I$: number of iterations

\State $P \leftarrow \{\}$ 
\For{$i=1$ \TO $n$} \Comment{\textbf{Sample initial HP configurations}}
    \State $hp \leftarrow$ Generate HP configuration
    \State $f_{hp} \leftarrow$ Evaluate performance of $hp$
    \State $P \leftarrow P \cup \{hp, f_{hp}\}$
\EndFor

\State $previous \leftarrow null$ 
\State $i = 1$
\While{$i < I$}
    \State $M \leftarrow$ Train a metamodel (ANN) using $P$
    
    \State $new_{hp} \leftarrow$ Sample the next HP configuration from a Gaussian distribution centered around the previously explored solution (or sample a random configuration if $previous = null$)
    
    \State $\widehat{f}_{new_{hp}} \leftarrow$ Predict the performance of $new_{hp}$ using the metamodel
    
    \If{$new_{hp}$ is predicted to be Pareto dominated}
        \State Select with certain probability $\alpha$ the configuration $\{new_{hp}\}$ to add to $P$ 
    \EndIf
    
    \If{$new_{hp}$ is accepted in $P$}
        \State $previous = \{new_{hp}, f_{new_{hp}}\}$
        \State $f_{hp} \leftarrow$ Evaluate performance of $hp$
        \State $P \leftarrow P \cup \{hp, f_{hp}\}$
        \State $i += 1$
    \EndIf
    
\EndWhile
    
\Return HP configurations in the Pareto front 
\end{algorithmic}
\end{megaalgorithm}  

\begin{megaalgorithm}
\caption{GP + GA Parsimony $g(n, I, N) = n+I+n+IN = 2n+I(N+1)$ }\label{alg:BO_GA}
\begin{algorithmic}[1]
\Require $n:$ initial design for BO, $I$: number of iterations, $N$: number of new configurations per iteration 

\Comment{\textbf{HPO using BO and training the ML with all the features}}
\State $P \leftarrow \{\}$ 
\For{$i=1$ \TO $n$} \Comment{\textbf{Sample initial HP configurations}}
    \State $hp \leftarrow$ Generate HP configuration
    \State $f_{hp} \leftarrow$ Evaluate performance of $hp$
    \State $P \leftarrow P \cup \{hp, f_{hp}\}$
\EndFor
\For{$i=1$ \TO $I$}
    \State $M \leftarrow$ Train a metamodel (GP) using $P$
    \State $new_{hp} \leftarrow$ Obtain a new HP configuration by optimizing an acquisition function using the metamodel predictions

    \State $f_{new_{hp}} \leftarrow$ Evaluate performance of $new_{hp}$
    \State $P \leftarrow P \cup \{new_{hp}, f_{new_{hp}}\}$
\EndFor

\Comment{\textbf{Feature selection using the best model HPs using a Genetic Algorithm}}
\State $T \leftarrow \{\}$ 
\For{$i=1$ \TO $n$} \Comment{\textbf{Create initial population}}
    \State $hpf \leftarrow$ Select HP configuration from $P$ and select a set of features of the ML problem
    \State $f_{hpf} \leftarrow$ Evaluate performance of $hpf$
    \State $T \leftarrow T \cup \{hpf, f_{hpf}\}$
\EndFor

\For{$i=1$ \TO $I$}
    
    \State $new_{hp1}, new_{hp2} \leftarrow$ Generate two HP configurations by applying genetic operators in two random HP configurations $hp1, hp2$ selected from $T$ 
    \State $f_{hp1}, f_{hp2} \leftarrow$ Evaluate performance of $new_{hp1}, new_{hp2}$
    \State $T \leftarrow T \cup \{\{new_{hp1}, f_{hp1}\}, \{new_{hp2}, f_{hp2}\}\}$
    \State Reduce $T$ to keep the same population size on each iteration
\EndFor
    
\Return HP configurations in the Pareto front 
\end{algorithmic}
\end{megaalgorithm}

\begin{megaalgorithm}
\caption{CMA-ES for MOO $g(n=0, I, N) = IN$ }\label{alg:CMA-ES}
\begin{algorithmic}[1]
\Require $I$: number of iterations, $N$: number of new configurations per iteration 

\State Create a multivariate normal distribution $\mathcal{N}(.)$ of $k$ hyperparameters configurations

\For{$i=1$ \TO $I$}
    \State $\{hp_1, \cdots, hp_N \} \leftarrow \mathcal{N}(.)$ \Comment{\textbf{Sample N candidates from} $\mathcal{N}(.)$}
    \State $\{f_{hp_1}, \cdots, f_{hp_N} \} \leftarrow$ Evaluate performance of each HP configuration
    
    \State Keep only the $l$ highest/lowest from $\{hp_1, \cdots, hp_N \}$ configurations using their non-dominating sort
    \State Update the multivariate normal distribution $\mathcal{N}(.)$ using the selected HP configurations
\EndFor

\Return HP configurations in the Pareto front 
\end{algorithmic}
\end{megaalgorithm}

\begin{megaalgorithm}
\caption{Multi-objective SA $g(n=0, I, N) = IN$ }\label{alg:MOSA}
\begin{algorithmic}[1]
\Require $I$: number of iterations, $N$: number of new configurations per iteration 

\For{$i=1$ \TO $I$} 
    \For{$q=1$ \TO $N$} 
        \State $hp \leftarrow $ Generate a new HP configuration from the neighborhood of the current best HP configuration $X$. A random configuration is used at the beginning of the optimization
        
        \State $f_{hp} \leftarrow$ Evaluate performance of $hp$
        
        \State Determine if $hp$ can be considered as the current best solution $X$ (acceptance rule defined to consider the dominance of $hp$ over $X$ and the configurations in an external archive of non-dominated solutions)
        
        \State Update, if needed, the external archive of non-dominated solutions
        
        \State Update the ``\textit{temperature}'' of the system 
        \EndFor
\EndFor

\Return HP configurations in the Pareto front 
\end{algorithmic}
\end{megaalgorithm}

\begin{megaalgorithm}
\caption{Nelder Mead algorithm with scalarized objectives $g(n, I, N) = n + 1 + I(N+1)$ }\label{alg:NMA}
\begin{algorithmic}[1]
\Require $n:$ initial solutions, $I$: number of iterations, $N$: number of new configurations per iteration 

\State $S \leftarrow \{\}$ 
\For{$i=1$ \TO $n+1$} \Comment{\textbf{Create initial Simplex}}
    \State $hp \leftarrow$ Generate an HP configuration
    \State $f_{hp} \leftarrow$ Evaluate performance of $hp$
    \State $S \leftarrow S \cup \{hp, f_{hp}\}$
\EndFor

\For{$i=1$ \TO $I$} 
    \State Sort vertices in $S$ in descending order
    \State $Sc \leftarrow$ Compute centroid vertex without the worst vertex
    \State $Sr \leftarrow$ Compute reflection of $Sc$
    \State $f_{Sr} \leftarrow$ Evaluate performance of $Sr$
    
    \If{$f_{Sr}$ is between the best and worst solution (excluding them)}
        \State Replace the worst solution in $S$ with $\{Sr, f_{sr}\}$ 
    \ElsIf{$f_{Sr}$ is better than the best solution}
        \State $Se \leftarrow$ Expand using $Sr$ and $Sc$
        \State $f_{Se} \leftarrow$ Evaluate performance of $Se$
        \State Replace the worst solution with $Se$ if this is better than $Sr$. Otherwise, use $Sr$
    \ElsIf{$f_{Sr}$ is worst than the current worst solution}
        \State $Scr \leftarrow$ Contract using $Sr$ and $Sc$
        \State $f_{Scr} \leftarrow$ Evaluate performance of $Scr$
        
        \If{$f_{Scr}$ is better than $f_{Sr}$}
            \State Replace the worst solution with $Scr$
        \Else \Comment{\textbf{Shrink toward the best solution}}
            \For{$j=2$ \TO $n+1$} 
                \State $Si \leftarrow$ shrink vertex $i$
                \State $f_{Si} \leftarrow$ Evaluate performance of $Si$
            \EndFor
        \EndIf
    \EndIf
\EndFor

\Return HP configurations in the Pareto front 
\end{algorithmic}
\end{megaalgorithm}

\begin{megaalgorithm}
\caption{MADE $g(n, I, N) = n + 2IN$ }\label{alg:MADE}
\begin{algorithmic}[1]
\Require $n:$ initial solutions, $I$: number of iterations, $N$: number of new configurations per iteration 

\State $P \leftarrow \{\}$ 
\For{$i=1$ \TO $n$} \Comment{\textbf{Create initial Population}}
    \State $hp \leftarrow$ Generate an HP configuration
    \State $f_{hp} \leftarrow$ Evaluate performance of $hp$
    \State $P \leftarrow P \cup \{hp, f_{hp}\}$
\EndFor

\For{$i=1$ \TO $I$}
    
    \For{$q=1$ \TO $N$} \Comment{\textbf{Generate new HP configurations}}
        \State $new_{hp} \leftarrow$ Generate a new HP configuration by applying mutation and crossover operators, using a random configuration $Rc$ selected from $P$
        
        \State $f_{new_{hp}} \leftarrow$ Evaluate performance of $new_{hp}$
        \State Replace $Rc$ with $new_{hp}$ if $new_{hp}$ dominates $Rc$
    \EndFor
    
    \state $S \leftarrow$ select non-dominated HP configurations
    \For{$q=1$ \TO $\mid S \mid$} \Comment{\textbf{Perform a chaotic local search around the solutions in the Pareto front}}
        
        \State $hp_{cls} \leftarrow$ Generate a new HP configuration around $S_q$
        \State $f_{hp_{cls}} \leftarrow$ Evaluate performance of $hp_{cls}$
        \State $P \leftarrow P \cup \{hp_{cls}, f_{hp_{cls}}\}$
    \EndFor
    
    \State $P \leftarrow$ Select HP configurations considering non-dominated and crowding distance sorting
    
\EndFor

\Return HP configurations in the Pareto front 
\end{algorithmic}
\end{megaalgorithm}

\end{appendices}


\newpage
\bibliography{references}


\end{document}